\documentclass[legalpaper, 10 pt, conference]{IEEEtran}  
\usepackage[letterpaper, left=0.75in, right=0.75in, bottom=0.75in, top=0.75in]{geometry}  
\usepackage{amsfonts}
\usepackage{array}
\usepackage{textcomp}
\usepackage{stfloats}
\usepackage{url}
\usepackage{verbatim}
\usepackage{graphicx}
\usepackage[utf8]{inputenc}
\usepackage{amssymb}
\usepackage{amsmath}
\usepackage{outlines}
\usepackage[ruled,vlined,linesnumbered]{algorithm2e}
\usepackage{tikz}
\usepackage{mathtools}
\usepackage{multirow}
\usepackage{caption}

\usepackage{subcaption}
\usepackage[style=ieee]{biblatex}
\bibliography{references}

\newcommand{\qt}[0]{\ensuremath{q_\theta}}
\newcommand*{\lossCE}{\ensuremath{\mathcal{L}_{\textsc{CE}}}}

\DeclarePairedDelimiterX{\infdivx}[2]{(}{)}{%
  #1\;\delimsize\|\;#2%
}

\newcommand{\kldiv}{\ensuremath{\textrm{KL}\infdivx}}

\DeclareMathOperator*{\E}{\mathbb{E}}

\DeclareMathOperator*{\argmin}{arg\,min}

\newcommand{\ofTheAppendix}{}

\IEEEoverridecommandlockouts  

\begin{document}

\title{AutoTransfer: Subject Transfer Learning with Censored Representations on Biosignals Data}

\author{
\IEEEauthorblockN{Niklas Smedemark-Margulies$^1$,}\thanks{$^1$Work done while NSM was on internship at Mitsubishi Electric Research Labs}
\IEEEauthorblockA{
Northeastern University \\
Boston, MA, USA \\
smedemark-margulie.n@northeastern.edu}
\and
\IEEEauthorblockN{Ye Wang,\quad \quad \quad Toshiaki Koike-Akino,}
\IEEEauthorblockA{
Mitsubishi Electric Research Labs. (MERL) \\
Cambridge, MA, USA \\
\{yewang, koike\}@merl.com}
\and
\IEEEauthorblockN{Deniz Erdo{\u{g}}mu{\c{s}}}
\IEEEauthorblockA{
Northeastern University \\
Boston, MA, USA \\
d.erdogmus@northeastern.edu}
}

\maketitle

\begin{abstract}
    We investigate a regularization framework for subject transfer learning in which we train an encoder and classifier to minimize classification loss, subject to a penalty measuring independence between the latent representation and the subject label. 
    We introduce three notions of independence and corresponding penalty terms using mutual information or divergence as a proxy for independence. 
    For each penalty term, we provide several concrete estimation algorithms, using analytic methods as well as neural critic functions.
    We propose a hands-off strategy for applying this diverse family of regularization schemes to a new dataset, which we call ``AutoTransfer''.
    We evaluate the performance of these individual regularization strategies under our AutoTransfer framework on EEG, EMG, and ECoG datasets, showing that these approaches can improve subject transfer learning for challenging real-world datasets.
\end{abstract}


\section{Introduction}

In this work, we investigate methods for transfer learning in the classification of biosignals data.
Previous work has established the difficulty of transfer learning for biosignals and even the issue of so-called ``negative transfer''~\cite{lin2017improving}, in which na\"{i}ve attempts to combine datasets from multiple subjects or sessions can paradoxically decrease model performance, due to differences in response statistics.
We address the problem of subject transfer by training models to be invariant to changes in a nuisance variable representing subject identifier (ID).
We examine previously established approaches and develop several new approaches based on recent work in mutual information estimation and generative modeling.
We evaluate these methods on a variety of electroencephalography (EEG), electromyography (EMG), and electrocorticography (ECoG) datasets, to demonstrate that these methods can improve generalization to unseen test subjects.
We also provide an automated hyperparameter search procedure for applying these methods to new datasets, which we call ``AutoTransfer''. 

Our basic approach to the transfer learning problem is to censor an encoder model, such that it learns a representation that is useful for the task while containing minimal information about changes in a nuisance variable (i.e., subject ID).
The motivation behind our approach is related to the information bottleneck method~\cite{info-bottleneck}, though with a key difference.
Whereas the information bottleneck and related methods seek to learn a useful and compressed representation from a supervised dataset without any additional information about nuisance variation, we explicitly use additional nuisance labels in order to draw conclusions about the types of variation in the data that should not affect our model's output.
Many transfer learning settings will have such nuisance labels readily available, and intuitively, the model should benefit from this additional source of supervision.
Our method ranked first place in the subject-transfer task of the NeurIPS BEETL challenge~\cite{beetl}.

The key contributions of this paper are three-fold:
\begin{itemize}
    \item We introduce a framework for subject transfer learning with nuisance-censored representations.
    \item We derive regularization penalties to enforce censoring via mutual information or divergence measures, and provide concrete estimation algorithms for these penalties using techniques including neural critic functions and analytic divergence estimates.
    \item We thoroughly evaluate these methods on challenging real-world subject-transfer datasets, showing that these methods improve generalization to unseen subject data.
\end{itemize}

\begin{figure}
    \centering
    \includegraphics[width=\linewidth]{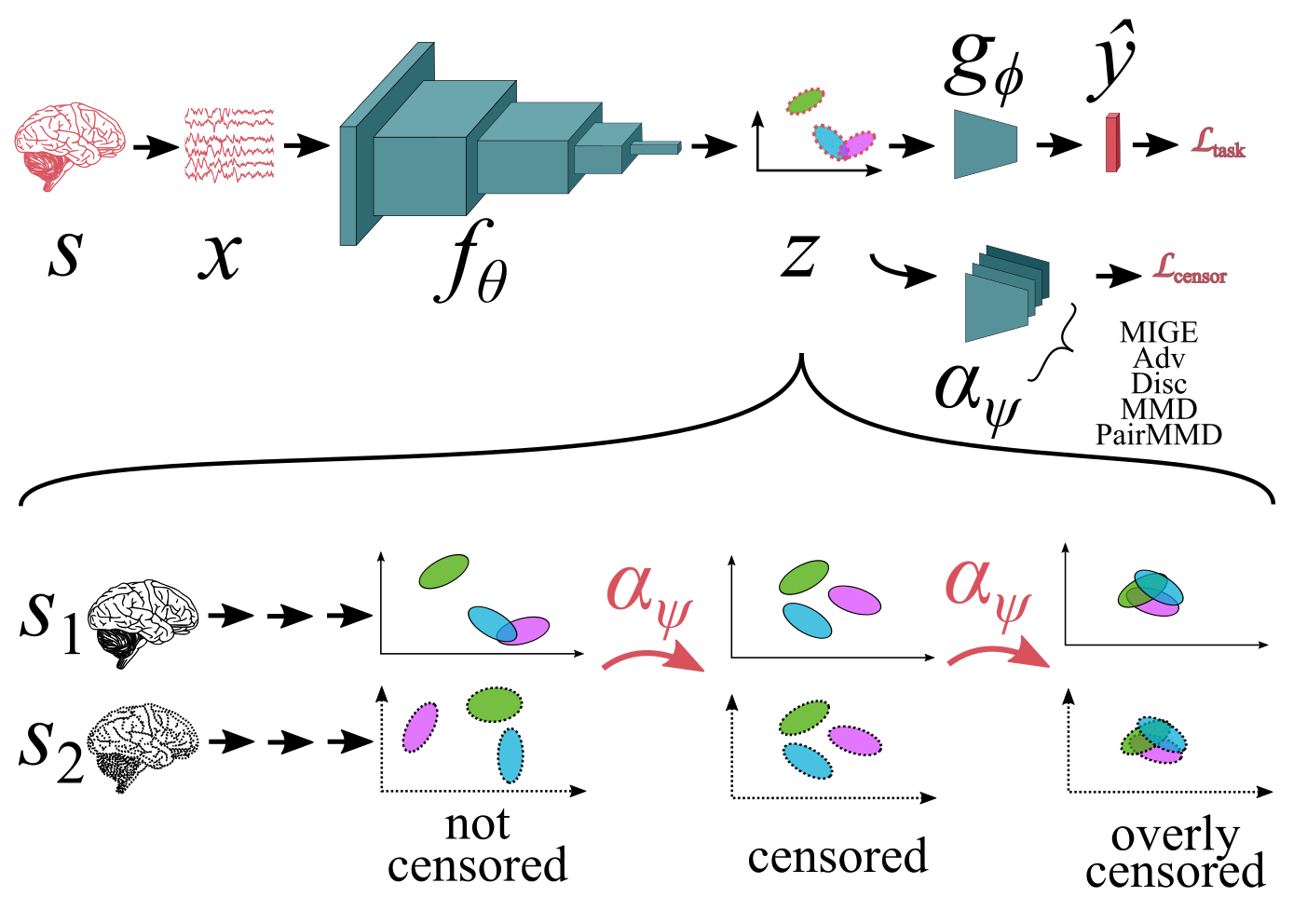}
    \caption{AutoTransfer pipeline for subject-invariant feature censoring in transfer learning. 
    Top: Model architecture. 
    Subject $s$ produces biosignals data $x$, which is mapped by encoder $f_\theta$ to latent code $z$. 
    Classifier $g_\phi$ gives class probabilities $\hat{y}$, resulting in task loss $\mathcal{L}_\mathrm{task}$.
    Censoring models $\alpha_\psi$ compute regularization penalty $\mathcal{L}_\mathrm{censor}$ to enforce independence. 
    Bottom: Regularization strategy. 
    Subjects $s_1, s_2$ are encoded, and their latent feature distributions are regularized.
    }
    \label{fig:schematic}
\end{figure}

\section{Learning Framework}
\label{sec:framework}

Consider a dataset $\{(x, y, s)\}_1^{N}$ consisting of $N$ triples of data $x \in \mathbb{R}^D$, discrete task labels $y \in \{1, \ldots, C\}$, and discrete nuisance labels $s \in \{1, \ldots, M\}$.
Let the generative model for the data distribution be defined as:
\begin{align}
    p_\mathrm{data}(x, y, s) = p(s) p(y | s) p(x | y, s).
\end{align}

Our transfer learning model consists of a parametric encoder $f_\theta(.): \mathbb{R}^D \to \mathbb{R}^K$ producing a $K$-dimensional latent representation $z = f_\theta(x)$, as well as a parametric task classifier $g_\phi(.): \mathbb{R}^K \to \mathbb{R}^C$.
We train this model to approximate the posterior distribution over labels given an input item $x$: $g_\phi(f_\theta(x)) \approx p_\mathrm{data}(y | x)$.
For a given choice of parameters $\theta$, our encoder model implies a conditional distribution $\qt(z|x)$ on features $z$ given the data $x$. 
We can then define other conditional distributions of the features as:
\begin{align}
    q_\theta(z | y, s) &= \int p(x | y, s) q_\theta(z | x) dx, \\
    q_\theta(z | y) &= \int p(x | y) q_\theta(z | x) dx, \\
    p(x | y) & = \int p(x | y, s) p(s | y) ds.
\end{align}

\subsection{Empirical Risk Minimization (ERM)}
In the standard ERM framework, we would seek to jointly learn parameters $\theta, \phi$ that minimize the expected classification risk, which we would approximate using an empirical average over our training data. 
For some specified loss function $\mathcal{L}$ and data distribution $p_\mathrm{data}(x, y, s)$, the risk is defined as the expected classification loss: 
\begin{align}
    R(\theta, \phi) = \E_{p_\mathrm{data}(x, y, s)} \left[ \mathcal{L}(g_\phi(f_\theta(x)), y) \right].
\end{align}

\subsection{Regularized ERM using Censoring} \label{sec:censoring_modes}
We consider a regularized form of the ERM framework, with an added penalty to enforce independence between the learned representation $z$ and the nuisance variable $s$, so that the classifier model $g_\phi$ can achieve similar performance across different subjects. 
This regularized learning framework is outlined in Fig.~\ref{fig:schematic}. 

We consider three overall notions of independence between latent representation and nuisance variable, which we refer to as ``censoring modes'':
\begin{enumerate}
    \item In ``marginal censoring'', we try to make representation $z$ \textit{marginally independent} of nuisance variable $s$: $z \perp s$.
    \item In ``conditional censoring'', we try to make $z$ \textit{conditionally independent} of $s$, given the task label $y$: $z \perp s | y$.
    \item In ``complementary censoring'', we partition the latent space into two halves $z = (z^{(1)},  z^{(2)})$, such that the first half $z^{(1)}$ is marginally independent of $s$, while maximizing the mutual information between the second half $z^{(2)}$ and $s$.
\end{enumerate}

Marginal censoring captures the simplest notion of a ``subject-independent representation''.
When the distribution of labels does not depend on the nuisance variable $p(y | s=s_1) = p(y | s=s_2)$, and the nuisance variable $s$ is therefore not useful for the downstream task, this marginal censoring approach will not conflict with the task objective.
However, there may naturally exist some correlation between $y$ and $s$ (i.e., subjects may perform the task differently); thus a representation $z$ that is trained to be useful for predicting the task labels $y$ may necessarily also be informative of $s$.
Conditional censoring accounts for this conflict between the task objective and censoring objective by allowing that $z$ contains some information about $s$, but no more than the amount already implied by the task label $y$.
Complementary censoring accounts for this conflict by requiring that one part of the representation $z$ is independent of the nuisance variable $s$, while allowing the other part to depend strongly on the nuisance variable. 

We capture these three censoring modes in a regularized ERM objective:
\begin{align}
    (\theta^*, \phi^*) = \argmin_{\theta, \phi} R(\theta, \phi) + \lambda\, \mathcal{L}_{\textrm{censor}}, \label{eqn:regularized-erm}
\end{align}
where $\mathcal{L}_{\textrm{censor}}$ is a penalty enforcing the desired independence. 
Table~\ref{tab:penalties} details the different forms for this penalty that we consider, and the estimation methods used for each.
\begin{table*}[t]
    \centering
    \caption{Conceptual forms for regularization penalty  $\mathcal{L}_{\textrm{censor}}$, and concrete estimation methods used for each penalty}
    \label{tab:penalties}
    \begin{tabular}{l l l l}
        \hline
        Censoring Mode & Desired Effect & Mutual Information form of $\mathcal{L}_{\textrm{censor}}$ & Divergence form of $\mathcal{L}_{\textrm{censor}}$ \\
        \hline
        Marginal & $z \perp s$ & $I(z;s)$ & $\mathcal{D}(\qt(z) || \qt(z|s))$ \\
        Conditional & $z \perp s | y$ & $I(z;s | y)$ & $\mathcal{D}(\qt(z|y) || \qt(z|s, y))$ \\
        Complementary & $z^{(1)} \perp s, \max I(z^{(2)},s)$ & $I(z^{(1)};s) - I(z^{(2)};s)$ & $\mathcal{D}(\qt(z^{(1)}) || \qt(z^{(1)}|s)) - \mathcal{D}(\qt(z^{(2)}) || \qt(z^{(2)}|s))$ \\
        \hline
        Estimation Methods & - & MIGE~\cite{mige}, Adversary~\cite{ozan-adversarial-inf}  & MMD/Pairwise MMD~\cite{mmd}, BEGAN Disc~\cite{began} \\ 
        \hline
    \end{tabular}
\end{table*}

\section{Estimation Techniques} \label{sec:estimation-techniques}
In this section, we derive several concrete methods for estimating the mutual information and divergence penalties used in the regularized objective functions outlined above.
Further details including pseudocode for the marginal, conditional, and complementary versions of each technique can be found starting in Appendix~\ref{sec:apx_adv}\ofTheAppendix{}.

\subsection{Mutual Information Estimation Methods}
We consider two ways to estimate the mutual information penalties required for the censoring objectives given above. 
First, we use an adversarial nuisance classifier, whose cross entropy loss provides a surrogate for the mutual information between $s$ and $z$ (see Section~\ref{sec:adv}). 
Second, we use Mutual Information Gradient Estimation (MIGE)~\cite{mige}, which uses score function estimators to compute the gradient of mutual information. 
We consider several kernel-based score function estimators (see Section~\ref{sec:mige}).

\subsubsection{Adversarial Censoring (Adv)} \label{sec:adv}

We consider minimizing the conditional mutual information between $z$ and $s$ given $y$ using an adversarial nuisance classifier model $\alpha$ with parameters $\psi$ that maps latent representations $z$ to a probability distribution over the nuisance variable $s$: $\alpha_\psi(.): \mathbb{R}^K \to \mathbb{R}^M$. 
Previous research~\cite{ozan-adversarial-inf, han2020disentangled1}
has established the technique of learning subject-invariant representations by training models in the presence of an adversarial subject classifier model.

Recall that for a given choice of encoder parameters $\theta$, we obtain representations $z = f_\theta(x)$ for each data point.
For a given choice of adversary parameters $\psi$ and encoder parameters $\theta$, computing the adversary's cross entropy loss $\mathcal{L}_{\textsc{CE, ADV}}(\theta, \psi) = \E_{p_\mathrm{data}(x, y, s)} [ - \log \alpha_\psi(s | z) ] $ gives an upper bound on the conditional entropy $H(s | z)$.
Noting that the mutual information can be decomposed as $I(z;s) = H(s) - H(s|z)$ and that the marginal entropy $H(s)$ is constant with respect to all model parameters, this gives a bound on the mutual information, which can be used as a surrogate objective for minimizing the mutual information:
\begin{align}
    I(z;s) \ge H(s) - \mathcal{L}_{\textsc{CE, ADV}}(\theta, \psi).
\end{align}
This bound will be tight for an adversary whose predicted distribution over subjects is close to the true posterior distribution; thus we can improve the quality of this surrogate objective by using a strong adversary model that is trained to convergence. 
See Appendix~\ref{sec:apx_adv}\ofTheAppendix{} for further details.

\subsubsection{Mutual Information Gradient Estimation (MIGE) Censoring} \label{sec:mige}

Given the difficulty of estimating mutual information in high dimensions, \textcite{mige} provide a method to estimate the \emph{gradient} of mutual information directly. 
This suffices for cases like ours, in which an objective function containing a mutual information term will be minimized by gradient descent. 
Appendix Section C of \textcite{ssge} derives a method for using a score function estimator to approximate the gradient of an entropy term. 
Appendix Sections A and B of \textcite{mige} show how to apply this idea for estimating the gradient of entropy terms to estimating the gradient of mutual information.

\paragraph{Score Function Estimation} \label{sec:score-estimators} 
The score function terms $\nabla_z \log \qt(z)$ and $\nabla_z \log \qt(z | s)$ required for MIGE penalties can be computed using any score function estimation method available. 
The original implementation by \textcite{mige} used the Spectral Stein Gradient Estimator (SSGE)~\cite{ssge}.
We explore other kernel-based score function estimation methods based on the work of \textcite{kscore}, who frame the problem of score function estimation as a regularized vector regression problem. 
See Appendix~\ref{sec:apx_score_estimation}\ofTheAppendix{} for further details about the estimators used and how their hyperparameters were set.

\subsection{Divergence Estimation Methods} \label{sec:divergence_methods} 
As outlined in Table~\ref{tab:penalties}, we also consider regularization penalties based on the divergence between two distributions. 
For marginal censoring, the definition of conditional probability tells us that the desired independence $z \perp s$ also implies that the distributions $\qt(z)$ and $\qt(z | s)$ are equivalent, or alternatively that the distributions $\qt(z|s_i)$ and $\qt(z|s_{j \ne i})$ are equivalent. 
Analogous divergences can be used for conditional or complementary censoring.
We provide three methods for censoring using divergence estimates; the first two are closely related, while the third is quite distinct.

The first two methods rely on a kernel-based estimate of the Maximum Mean Discrepancy (MMD)~\cite{mmd}, which provides a numerical measure of distance between two distributions.
The MMD between two distributions is $0$ when the distributions are equivalent.
In Section~\ref{sec:mmd}, we use MMD as a surrogate for the divergence between $\qt(z)$ and $\qt(z|s)$, which we refer to as simply the ``MMD'' censoring approach.
In Section~\ref{sec:pairmmd}, we use MMD as a surrogate for the divergence between $\qt(z|s_i)$ and $\qt(z|s_{j \ne i})$, which we refer to as the ``Pairwise MMD'' censoring approach.

In the third method, we use a neural discriminator model based on Boundary Equilibrium Generative Adversarial Networks (BEGAN)~\cite{began}.
In the original work, this discriminator provides a surrogate measure of the divergence between real and generated data distributions.
In our work, we use the discriminator to provide a measure of the divergence between $\qt(s)$ and $\qt(z | s)$, which allows us to reduce the dependence of $z$ on $s$.
See Section~\ref{sec:disc} for further details.

\subsubsection{MMD Censoring} \label{sec:mmd}
The MMD~\cite{mmd} provides a desirable measure of divergence between distributions because it makes no assumptions about the parametric form of the distributions being measured, and because it can be approximated efficiently with a kernel estimator, given a batch of samples from each distribution.
The MMD is an integral probability metric, describing the divergence between two distributions $p(\cdot)$ and $q(\cdot)$ as the difference between the expected value of a test function $f \in \mathcal{F}$ under each distribution, for some worst-case $f$ from a class of functions $\mathcal{F}$:
\begin{align}
    \! \mathsf{MMD}(\mathcal{F}, p, q) = \sup_{f \in \mathcal{F}} \Big(
    \E_{x \sim p(x)} \left[ f(x) \right] - \E_{y \sim p(y)} \left[ f(y) \right]
    \Big).
\end{align}
\textcite{mmd} derive a kernel estimate of the MMD using a radial basis function kernel (see details in Appendix~\ref{sec:apx_mmd}\ofTheAppendix{}). 
We use their empirical estimate, with a kernel length scale set by the median heuristic~\cite{sriperumbudur2009kernel} as we do for a subset of MIGE experiments. 

\subsubsection{Pairwise MMD (PairMMD) Censoring}
\label{sec:pairmmd}

Computing the MMD between $\qt(z)$ and $\qt(z | s)$ provides us a quantitative measure of the dependence between $z$ and $s$, and by minimizing it we can enforce the indepences we desire.
We can similarly measure the divergence between $\qt(z | s_i)$ and $\qt(z | s_{j \ne i})$ to enforce these independences.
To compute an overall penalty using this ``pairwise'' approach, we consider all combinations of $\binom{M}{2}$ distinct values of the nuisance variable, and compute an average over these individual terms.
Since computing the full quadratic set results in a potentially large overhead, we consider two approximations by selecting a random subset of terms to compute as below.

First, we consider using a parameter $b \in [0, 1]$ controlling a Bernoulli distribution to select a random subset of all possible pairs of $s_i, s_j$ for $i\neq j$, which we call a ``Bernoulli'' subset selection.
Second, we consider using an integer $d \in \{1, \ldots, M\}$ controlling the number of nuisance values included, and compute a term for all combinations within this subset, which we call a ``clique'' subset selection.
These two selection procedures are described in more detail in Appendix~\ref{sec:apx_pairmmd}\ofTheAppendix{}.

\subsubsection{BEGAN Discriminator Censoring}
\label{sec:disc}

BEGAN~\cite{began} uses an adversarial training scheme to learn a generative model. 
A generator network $G$ tries to approximately map samples from a unit Gaussian distribution in its latent space to samples from the target data distribution, while a discriminator network $D$ tries to distinguish real and fake data samples, as in a standard GAN setup~\cite{goodfellow2014generative}.
This model uses an autoencoder as the discriminator to compute a lower bound on the Wasserstein-1 distance between the distribution of its autoencoder loss on real and generated data. 
In other words, the discriminator separates the two distributions by learning an autoencoder map that works well only for the ``true'' data distribution; the generator tries to produce data that matches the ``true'' data distribution and is well-preserved by this autoencoder map.
The training is further stabilized by introducing a trade-off parameter to adaptively scale the magnitude of the discriminator's two loss terms.

The role of the discriminator in this original setup is to provide a surrogate objective so that the generator can bring two distributions (the true data distribution $p_\mathrm{data}(x)$, and its own generated distribution $p_{G}(x)$) closer together.
We use their method to provide a signal that allows our encoder model $f_\theta$ to approximately minimize the divergence terms.

\section{Experiments} \label{sec:experiments}

In order to evaluate the proposed regularization approaches described in~(\ref{eqn:regularized-erm}) and Table~\ref{tab:penalties}, we perform experiments with several challenging real-world datasets. 
For each dataset, we explore all of the censoring estimation procedures described above.
For detailed pseudocode, see Algorithms 
\ref{alg:adv_marginal} through
\ref{alg:disc_complementary},
detailed in Appendix~\ref{sec:apx_pairmmd}\ofTheAppendix{}.
We first search for promising hyperparameter ranges, then evaluate the most promising subset of hyperparameters using $k$-fold cross-validation and evaluate our AutoTransfer method on the resulting collection of models. 

\subsection{Datasets} \label{sec:datasets}

We use a diverse set of physiological datasets: EEG (rapid serial visual presentation, RSVP~\cite{rsvp-dataset}; error-related potentials, ErrP~\cite{errp-dataset}), EMG (American Sign Language, ASL~\cite{asl-dataset}), and ECoG (facial recognition, EcogFacesBasic~\cite{ecog-faces-basic}).
To standardize the comparison across datasets, all data were preprocessed by z-scoring each channel of each trial. 
Additional feature engineering could improve absolute performance, 
though such techniques are orthogonal to the focus of the present study.
For further dataset details, see Appendix~\ref{sec:apx_datasets}\ofTheAppendix{}.

\begin{table*}[t]
    \centering
    \caption{Censoring hyperparameters explored in AutoTransfer}
    \label{tab:hyperparams}
    \begin{tabular}{lll}
        \hline
        Censoring Method & Parameter & Values Explored \\
        \hline
        Adv, MIGE, BEGAN Disc & Regularization Coefficient $\lambda$ & $1, 0.3, 0.1, 0.03, 0.01$\\
        MMD, PairMMD & Regularization Coefficient $\lambda$ & $1, 3, 10, 30, 100$ \\
        MIGE & Score Function Estimator $F_\mathrm{score}$ & SSGE~\cite{ssge}, MIGE, $\nu$-Method, Tikhonov, Stein~\cite{kscore} \\
        MIGE & Score Function Estimator Regularization $\gamma$ &  $0.01, 0.001, 0.0001$ \\
        MIGE & Adaptive Length-scale Method & Median, t-SNE-style \\
        \hline
    \end{tabular}
\end{table*}

\subsection{Network Architectures and Training}

The neural network architectures for our feature extractor network $f_\theta$ is based on EEGNet~\cite{eegnet}.
The classifier network $g_\phi$ consists of a single linear layer with softmax activation.

Models are trained to maximize balanced accuracy by using weighted cross entropy; for class $k$ with $N_k$ examples in the training set, the unnormalized weight $\tilde{w}_k$ is set as the inverse of the class proportion $\tilde{w}_k = \sum_i N_i / N_k$, and then the sum of weights is normalized to one, $w_k = \tilde{w}_k / \sum_i \tilde{w}_i$. 

In each training fold, we designate one held-out subject for validation and one for test. 
The validation subject is used for model selection, as well as for early stopping. 
Models are trained for a maximum $500$ epochs using the AdamW optimizer~\cite{adamw}. 
We begin with a learning rate of $\alpha_1 = 10^{-3}$, and decay the learning rate by the inverse square-root of the epoch number, such that at epoch $t$, we use a learning rate of $\alpha_t = \alpha_1 / \sqrt{t}$. 
The epoch of minimum validation loss is then evaluated on the test subject.

\begin{figure*}[!th]
    \centering
    \begin{subfigure}[h]{0.49\textwidth}
    \includegraphics[width=\textwidth]{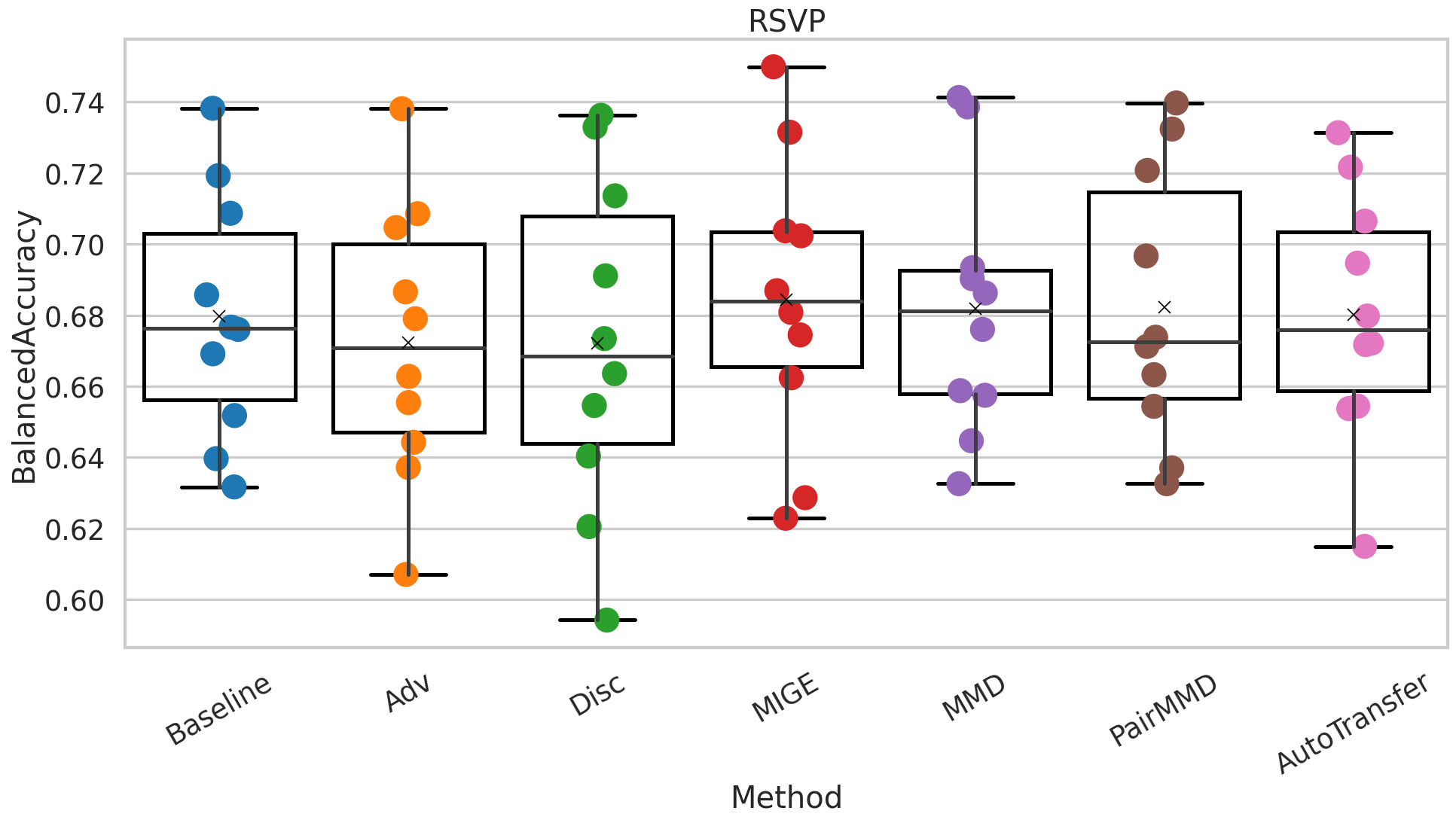}
    \includegraphics[width=\textwidth]{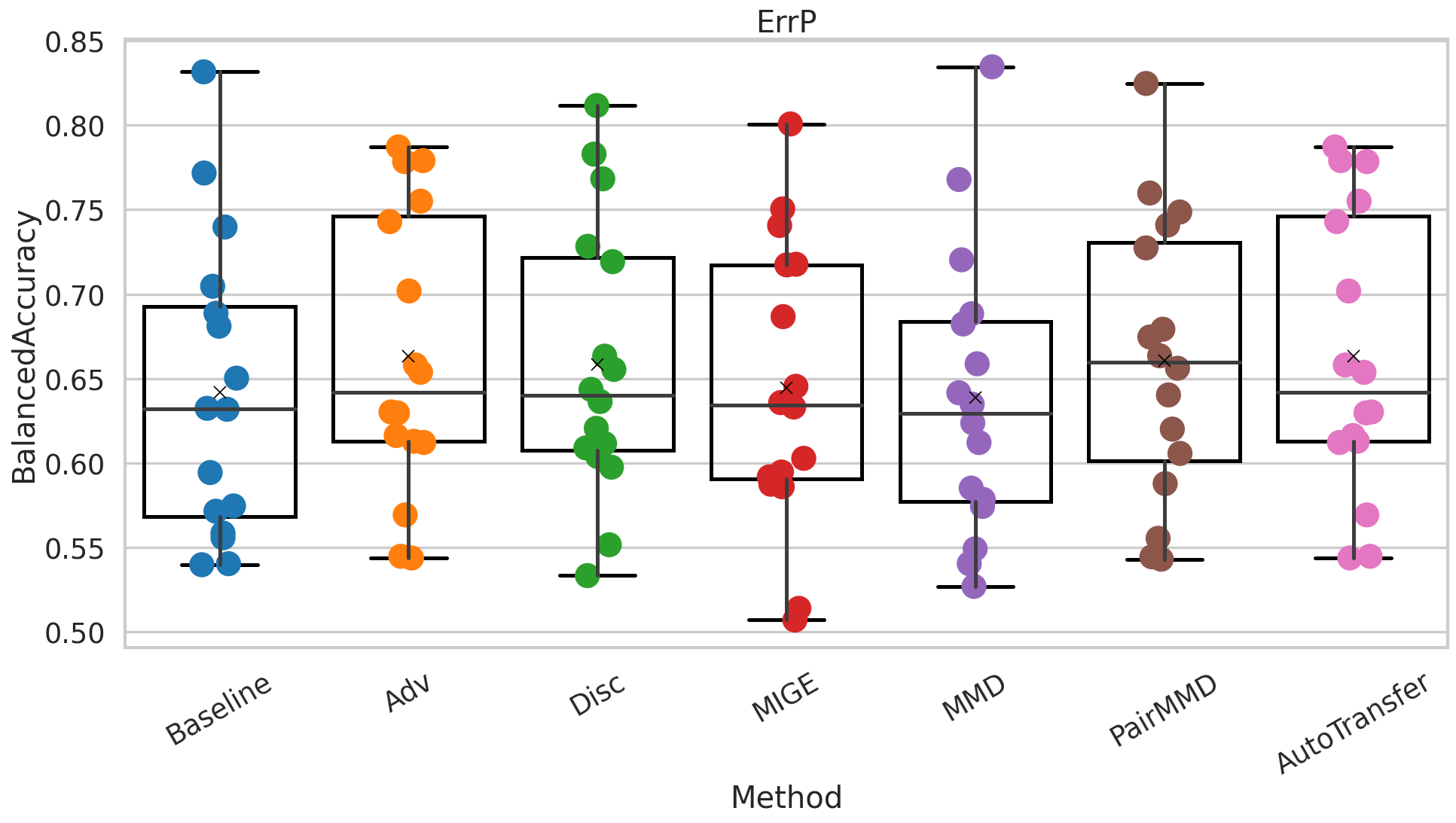}
    \includegraphics[width=\textwidth]{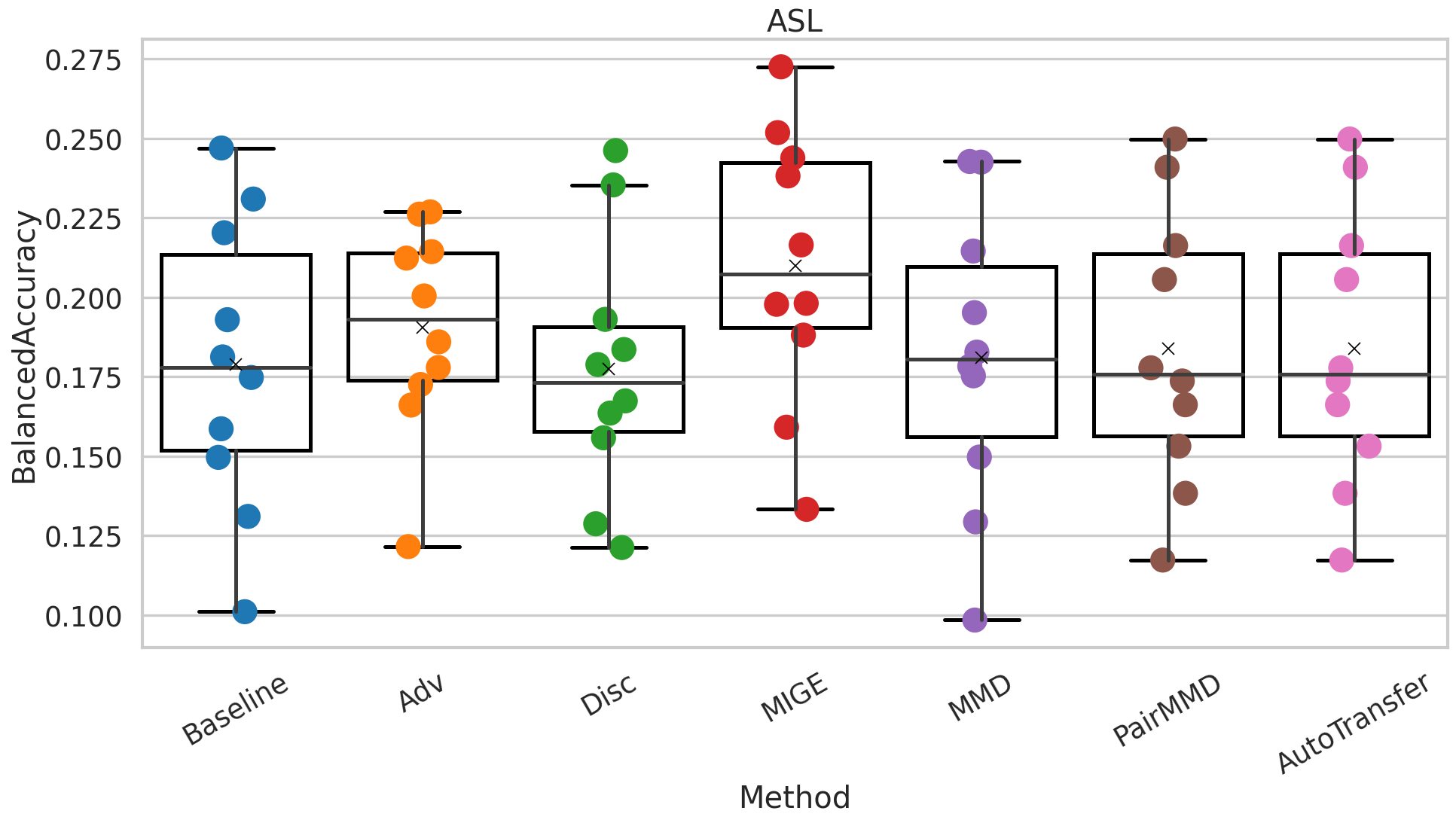}
    \includegraphics[width=\textwidth]{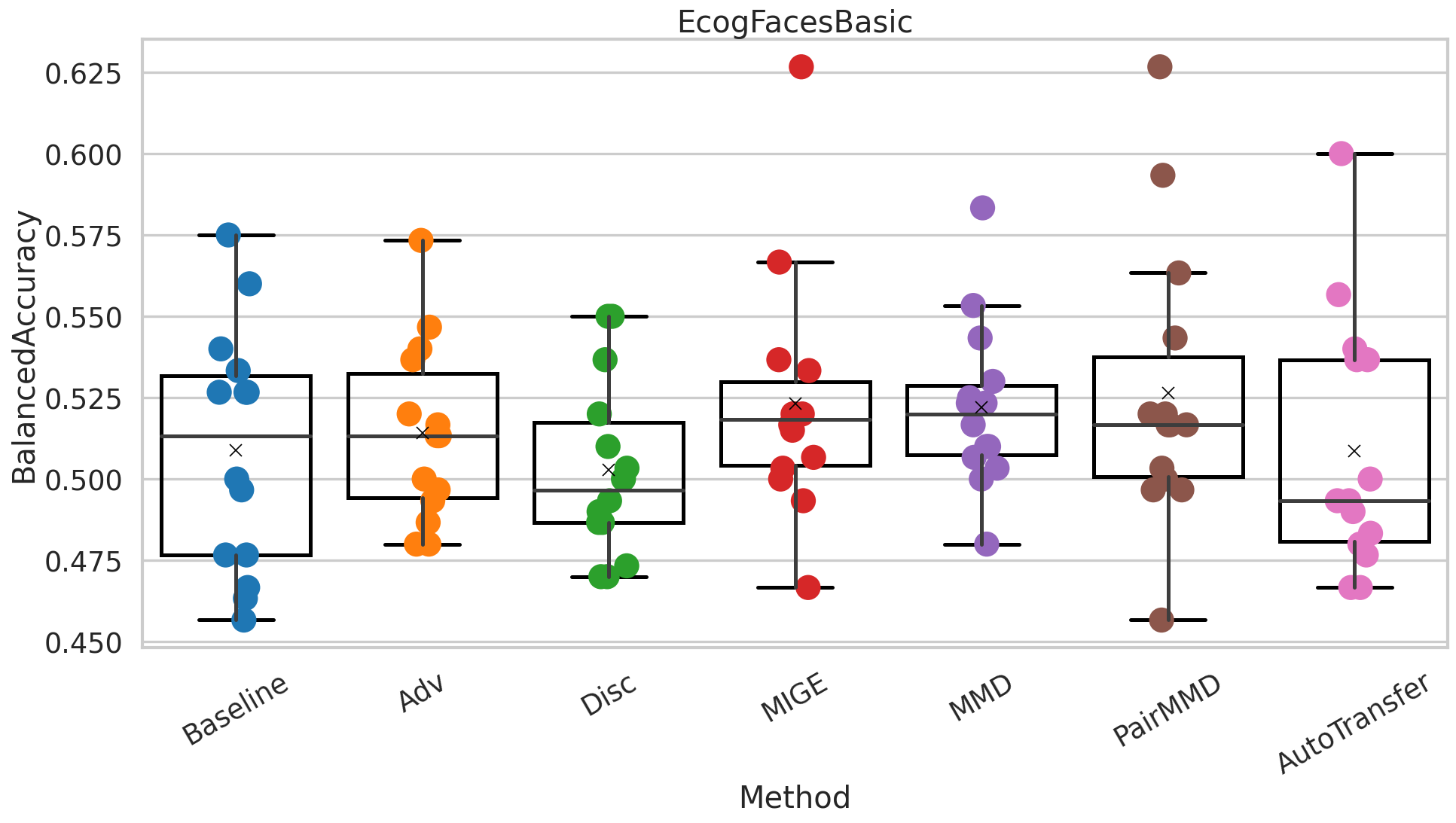}
    \end{subfigure}
    \hfill
    \begin{subfigure}[h]{0.49\textwidth}
    \includegraphics[width=\textwidth]{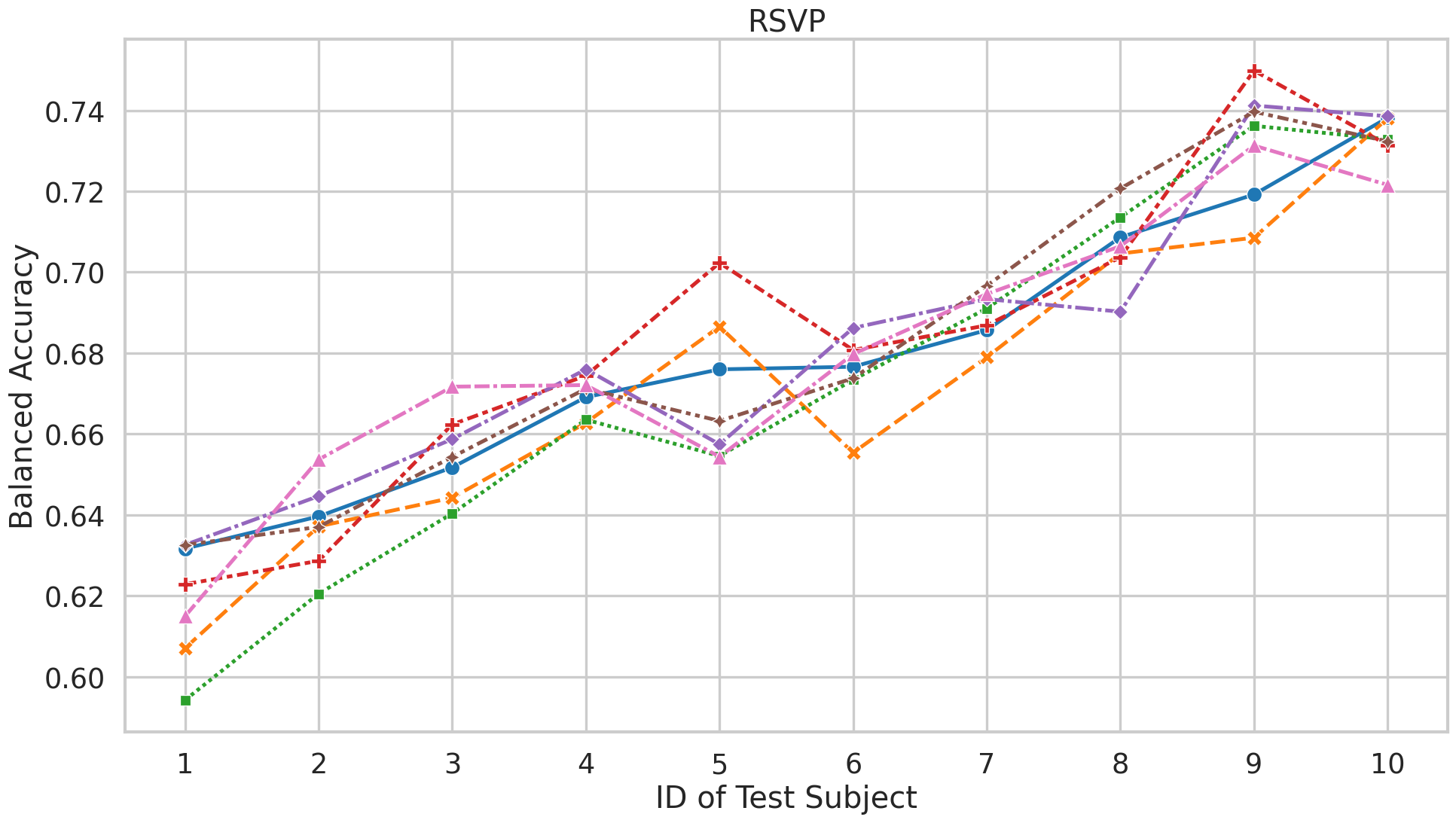}
    \includegraphics[width=\textwidth]{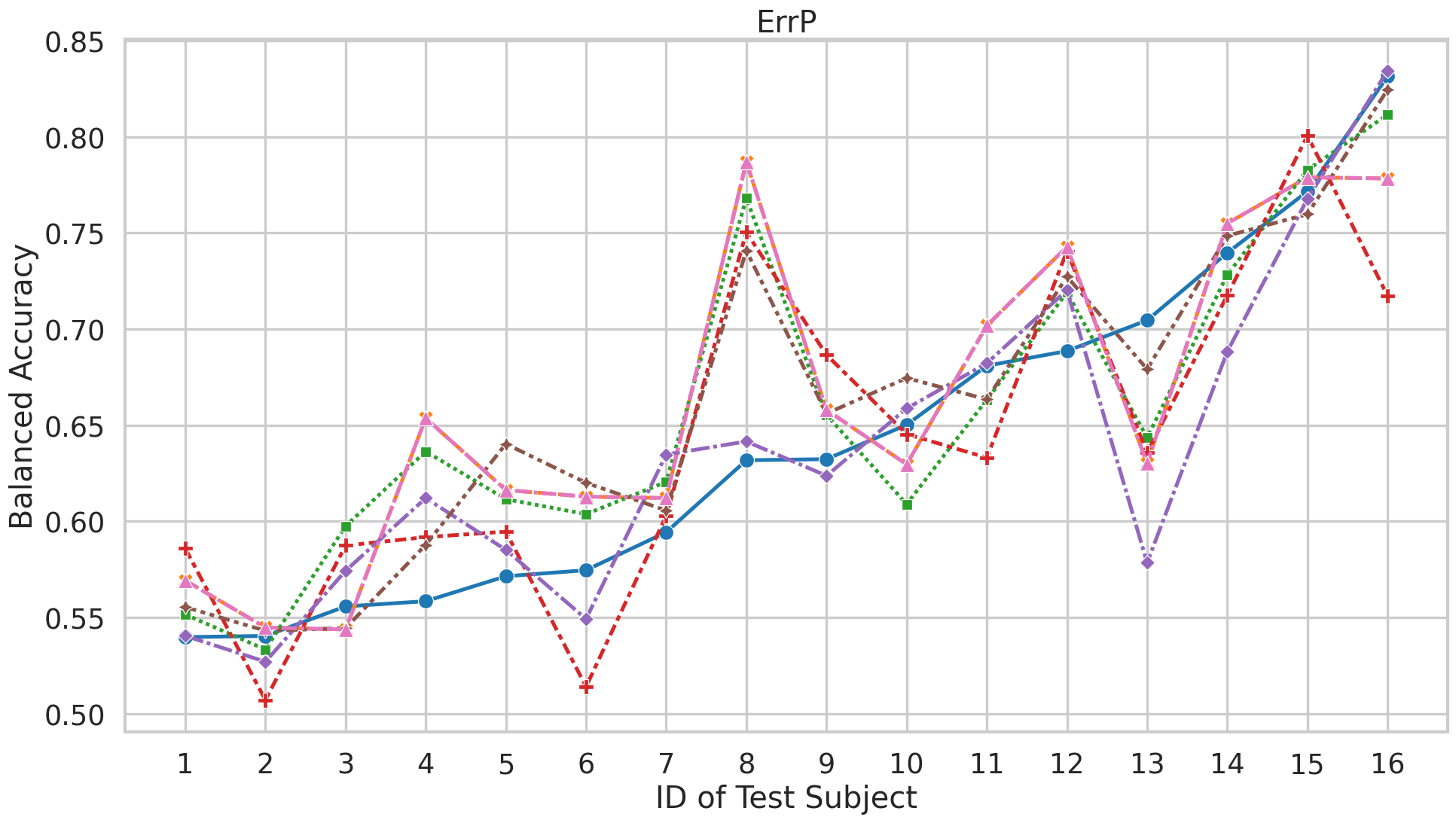}
    \includegraphics[width=\textwidth]{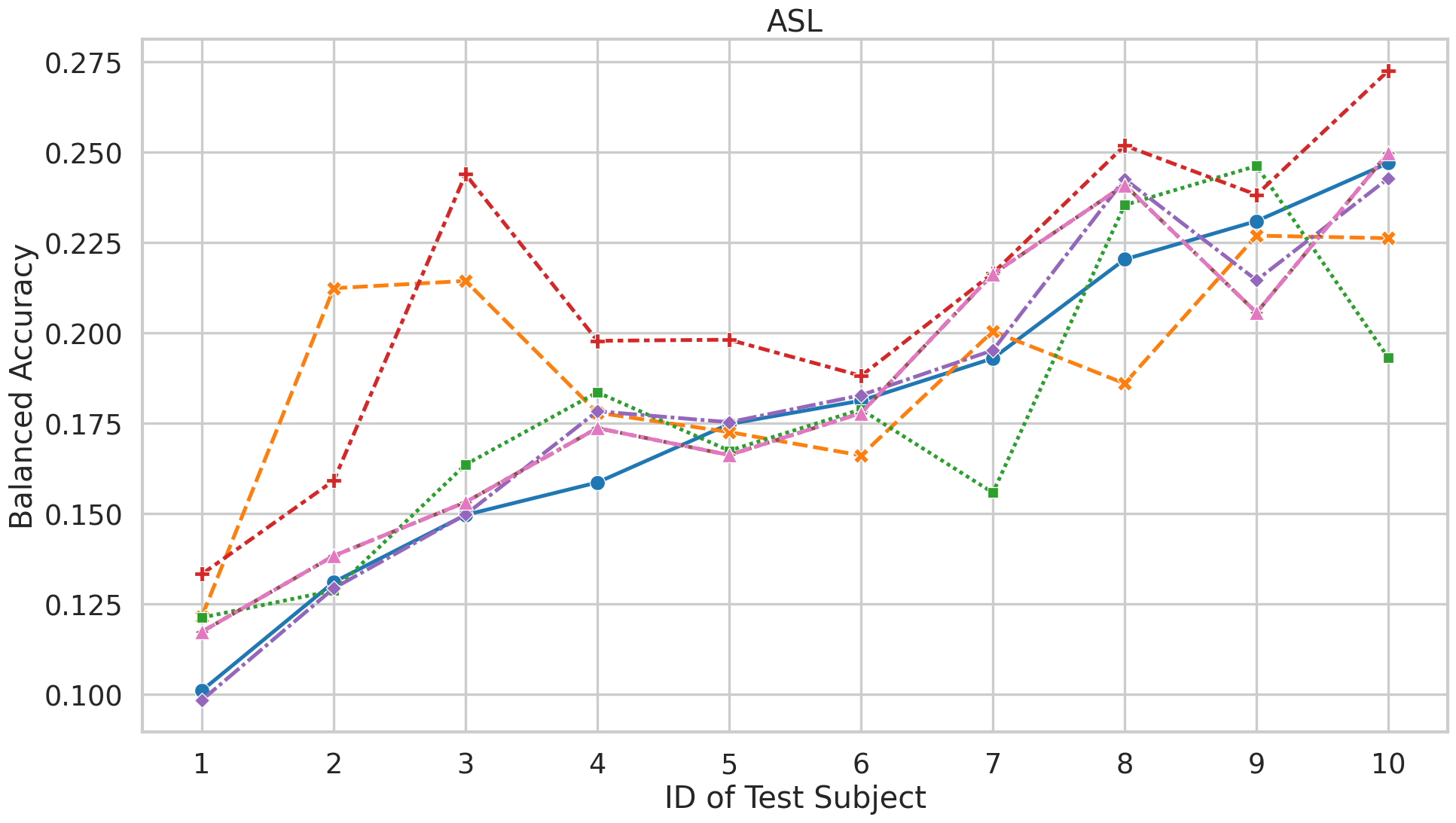}
    \includegraphics[width=\textwidth]{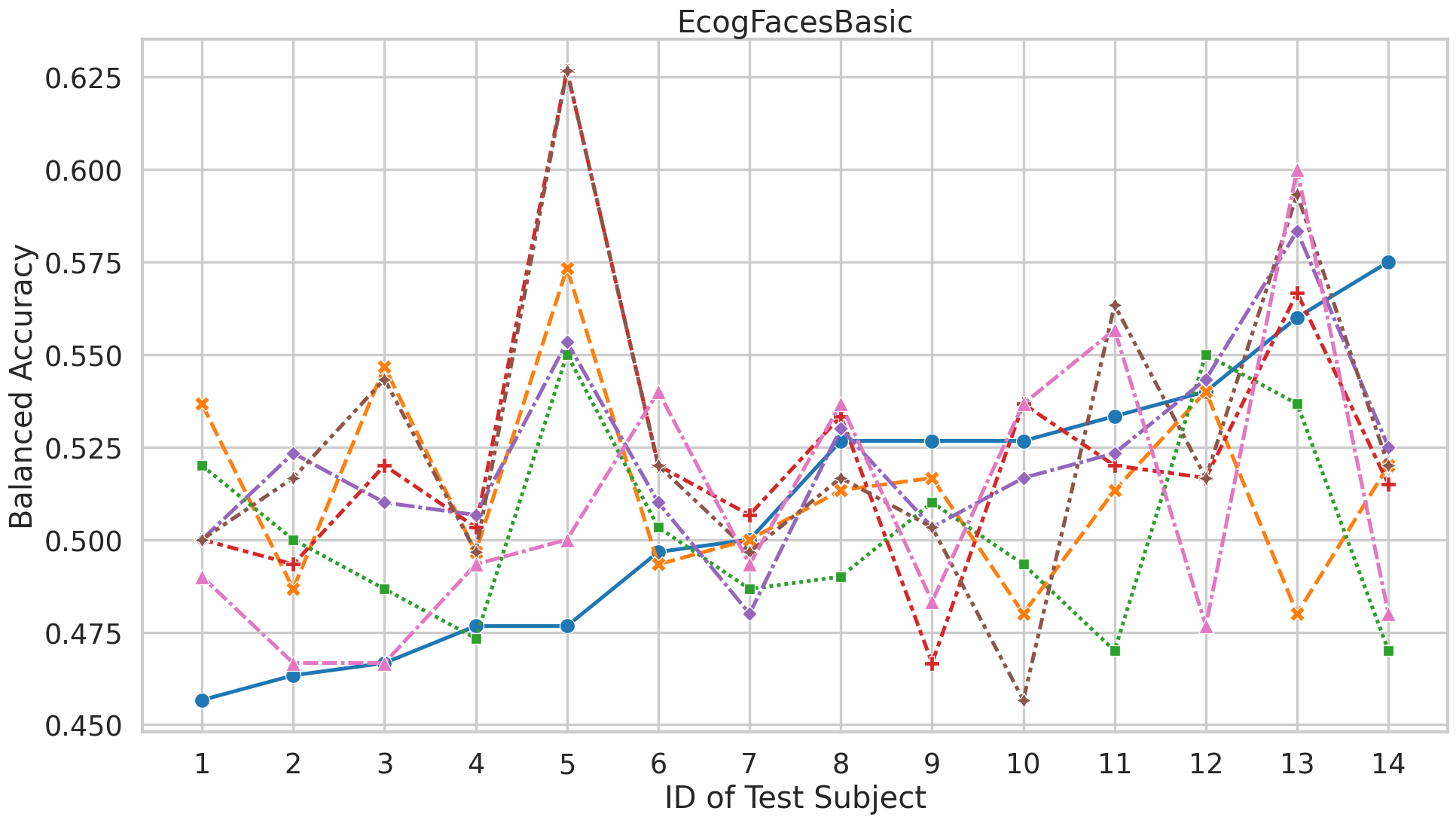}
    \end{subfigure}
    \caption{
    Subject transfer balanced accuracy. 
    Left: Test score from each fold, black `x' indicates mean. 
    Right: Accuracy vs test subject, sorted by baseline performance. 
    Same colors for left and right.
    Censoring improves subject transfer, especially in subjects whose transfer performance is initially low. The best single censoring method is dataset dependent, while AutoTransfer consistently performs well and often matches the best method.
    See Sec.~\ref{sec:datasets} for dataset information.
    }
    \label{fig:results}
\end{figure*}

\subsection{Hyperparameter Tuning} \label{sec:hyperparam-tuning}

For each dataset under consideration, we tune several key hyperparameters for each of our estimation methods. 
We use the same data split for all settings explored; one subject is kept for validation, and another is kept for testing.
We then select the best $3$ settings for each of the five methods discussed above according to balanced validation accuracy.
For each method, we vary the Lagrange multiplier coefficient $\lambda$. 
For the MIGE censoring, we also vary the score function estimator between the default SSGE, and three alternative kernel-based estimators discussed in Section~\ref{sec:score-estimators}. 
For these three alternative score function estimators, we vary their own internal regularization parameter $\gamma$,
and vary the method of setting the kernel length scale as discussed in Appendix~\ref{sec:apx_adaptive_lengthscale}\ofTheAppendix{}. 
Table~\ref{tab:hyperparams} summarizes the range of hyperparameters explored for each method.
Note that this hyperparameter search for both finite descrete values and contineous values can be fully automated by, e.g., Bayesian optimization in an AutoML framework~\cite{akiba2019optuna}.

\subsection{Cross-Validation} \label{sec:cross-validation}

We select the best $3$ combinations of hyperparameter settings according to validation accuracy for each method for further examination by $k$-fold cross-validation.
Specifically, for a dataset with $M$ subjects, the cross-subject validation gives us a collection of $M$ test accuracies, which we can visualize as a distribution of model performance.
We then apply our AutoTransfer procedure; for a given dataset, we select the method whose 25\textsuperscript{th}-percentile validation accuracy is highest. 
We select methods based on lower quartile performance as a way to avoid overfitting to the validation subjects.

Fig.~\ref{fig:results} shows the results of these cross-validation experiments.
On the left, each dot represents the transfer performance when a certain subject is used as the test set. 
The most striking observation is the wide variation in subject transfer performance; this is the heart of the difficulty in subject transfer learning. 
For each dataset, we observe that at least one of the estimation methods provides an improvement to the interquartile range, despite the presence of outlier subjects. 
Although the most popular censoring method based on adversarial training works well for most datasets, other censoring approach such as MIGE censoring can outperform it for some datasets.
This suggests us that exploring different censoring methods is of great importance depending on datasets.
On the right, each x-axis position represents a single test subject, and they are sorted according to their baseline transfer performance. 
Here we can see that our regularization penalties offer a strong benefit for some subjects, especially those whose baseline transfer accuracy is relatively worse.

\section{Discussion}

We addressed the problem of subject transfer learning for biosignals datasets by using a regularized learning framework to enforce one of several possible notions of independence, which we refer to as censoring modes. 
We derived estimation algorithms for each of the proposed regularization penalties.
We evaluated these estimation algorithms on a variety of challenging real-world datasets including EEG, EMG, and ECoG, and found that these methods can offer significant improvement, especially for subjects originally near the lower quartile of transfer performance.
Finally, we provided an automated end-to-end procedure for exploring and selecting a censoring method on a new dataset, which we call AutoTransfer.
In our cross-validation experiments, AutoTransfer consistently offers an improvement over baseline transfer performance, though it may be below the maximal single-method performance due to the inherent inter-subject variability in these tasks.
Note that we considered the case of discrete nuisance variables (subject ID), though most techniques are readily applicable to the case of continuous-valued nuisance variables.

\paragraph*{Future Work} 
Our proposed approach is designed to improve subject transfer performance without making use of test-time adaptation or information about the statistics of the transfer subject's data. 
In order to construct a holistic transfer learning system, future work may therefore combine our training-time regularization with other test-time adaptation strategies such as few-shot learning,
parameter prediction~\cite{bertinetto2016learning, requeima2019fast, perez2018film}, 
data augmentation~\cite{sohn2020fixmatch, cubuk2019autoaugment},
and self-supervision~\cite{zhang2017mixup, tarvainen2017mean, miyato2018virtual, berthelot2019mixmatch}.

Furthermore, our methods are compatible with a number of other standard machine learning techniques that may improve absolute model performance. 
For example, feature engineering and hand-tuned feature extraction provide a way to strictly enforce prior knowledge about signal characteristics; when collecting biosignals datasets, this might include information such as sensor artifacts and signal frequency ranges.
Since we propose a large family of estimation algorithms, our work may naturally benefit from model ensembling techniques, though the challenge in this context is to handle the overfitting issues that are inherent to these challenging subject transfer datasets.
This large family of estimation algorithms also comes with a large set of model hyperparameters to consider; it is likely that further hyperparameter search may offer additional improvement in the final model performance. 
Finally, we note that it may be possible to use a decoder model with a reconstruction loss term as part of a method for enforcing the regularization strategies.

\printbibliography

\clearpage
\onecolumn

\appendix

Here we give detailed pseudocode and complete derivations for the marginal, conditional, and complementary versions of each estimation technique used for computing a censoring penalty.

\subsection{Mutual Information Estimation Methods}

As mentioned previously, one of the high-level ways to enforce independence between the representation $z$ and the subject label $s$ is to reduce some measure of mutual information between these variables. In the marginal censoring case, we seek to minimize $I(z;s)$. 
In the conditional case, we seek to minimize $I(z;s|y)$.
In the complementary case, we seek to minimize $I(z^{(1)};s)$ and maximize $I(z^{(2)};s)$.

\subsubsection{Adversarial Censoring} 
\label{sec:apx_adv}

Recall that we can use the cross entropy loss of an adversarial classifier model as a surrogate objective for minimizing the mutual information between representation and subject label $I(z;s)$ needed for marginal censoring, as well as the conditional mutual information $I(z;s|y)$ or the two terms $I(z^{(1)};s)$ and $I(z^{(2)};s)$ used for complementary censoring.

\paragraph{Marginal Mutual Information}
We use an alternating optimization scheme, training the adversary's parameters $\phi$ to minimize a standard cross entropy loss for predicting the subject label.
This can be seen as minimizing an upper bound on the conditional entropy $H(s | z)$ as follows. Here, let $q_\psi(s|z)$ represent the distribution of labels predicted by the adversary $\alpha_\psi$ when given input $z$.
\begin{align}
    \psi^* & = \argmin_\psi \underbrace{\E_{p_{data}(x, y, s)} [ - \log \alpha_\psi (s | \underbrace{z}_{=f_\theta(x)}) ]}_{\text{cross-entropy } \mathcal{L}_{\textsc{CE}}(\theta, \psi)} \\
    \mathcal{L}_{\textsc{CE}}(\theta, \psi) 
    & = \E_{p(s, z)} \left[ - \log p(s | z) \right] + \E_{p(s, z)} \left[ \log \frac{p(s | z)}{q_\psi(s | z)} \right] \\
    & = \E_{p(s, z)} \left[ - \log p(s | z) \right] + \E_{p(s | z)} \left[ \log \frac{p(s | z)}{q_\psi(s | z)} \right] \\
    & = H(s | z) + \underbrace{ \kldiv{p(s | z)}{q_\psi(s | z)}}_{\ge 0} \\
    \mathcal{L}_{\textsc{CE}}(\theta, \psi) & \ge  H(s | z)
\end{align}

For a given choice of adversary parameters $\psi$, computing the cross-entropy loss gives an upper bound on the conditional entropy $H(s | z)$;
note that the mutual information can be decomposed as $I(z;s) = H(s) - H(s | z)$, and that the entropy of the nuisance variable $H(s)$ is constant with respect to our parameters $\theta, \psi, \phi$. 
Thus, we can substitute the adversary's cross entropy loss to obtain a lower bound on the mutual information:
\begin{align}
    I(z;s) & = H(s) - H(s | z) \\
    & \ge H(s) - \lossCE(\theta, \psi)
\end{align}

Finally, we can use this bound as a surrogate for the regularization term in our training objective for the encoder and classifier models.
Since $H(s)$ is constant with respect to all parameters, we can ignore it during minimization.
Note that this bound will be tight when the adversary's predicted distribution $q_\phi(s | z)$ is close to the true posterior $p(s | z)$ (such that the KL divergence between them is small).
\begin{align}
    \theta^*, \phi^*
    & = \argmin_\theta R(\theta, \phi) + \lambda I(z; s) \\
    & \approx \argmin_\theta R(\theta, \phi) - \lambda \lossCE(\theta, \psi)
\end{align}
For simplicity of notation in pseudocode, let $\mathcal{L}_{\textsc{CE}}(q_\psi(s_i | z_i), s_i)$ represent the cross-entropy loss between the true one-hot distribution represented by the label $s_i$, and the predicted label distribution $q_\psi(s_i | z_i)$ produced by the adversary network. Note that this notation omits the fact that the representation $z_i$ depends on the encoder's parameters $\theta$. 
This results in the censoring procedure in Alg.~\ref{alg:adv_marginal}.

\begin{algorithm}[h]
\DontPrintSemicolon
\KwIn{Samples $\{(x_i, y_i, s_i)\}_{i=1}^N$, Encoder $f_\theta$, Adversarial Classifier $\alpha_\psi$}
\KwOut{Mutual Information penalty}
$\mathcal{L}_{\textrm{total}} \gets 0$ \;
\For{$i$ in $1 \ldots N$}{
    $z_i \gets f_\theta(x_i)$ \tcp*[r]{Encode raw data}
    $q_\psi(s_i | z_i) \gets \alpha_\psi(z_i)$ \tcp*[r]{Predict subj}
    add $\lossCE(q_\psi(s_i | z_i), s_i)$ to $\mathcal{L}_{\textrm{total}} $ \;
}
\Return $\mathcal{L}_{\textrm{total}}$ \;
\caption{Marginal Adversarial Censoring}
\label{alg:adv_marginal}
\end{algorithm}


\paragraph{Conditional Mutual Information} We can also use an adversarial classifier model to provide an approximation of the conditional mutual information $I(z;s | y)$.
In this case, we train the adversary to predict the nuisance variable $s$ given both the representation $z$ and the task label $y$ by minimizing the cross-entropy loss:
\begin{align}
    \psi^* & = \argmin_\psi \underbrace{\E_{p_{data}(x,y,s)} \left[ - \log \alpha_\phi(s | z, y) \right]}_{\text{cross-entropy} \mathcal{L}_{\textsc{CE}}(\theta, \psi)} \\
    \mathcal{L}_{\textsc{CE}}(\theta, \psi) 
    & = \E_{p_{data}(x, y, s)} [ - \log p(s | z, y)]  + \E_{p_{data}(s | z, y)} \left[ \log \frac{p(s | z, y)}{\alpha_\psi(s | z, y)} \right] \\
    & = H(s | z, y) + \kldiv{p(s | z, y)}{\alpha_\psi(s | z, y)} \\
    \mathcal{L}_{\textsc{CE}}(\theta, \psi) & \ge H(s | z, y)
\end{align}

As before, note that the conditional mutual information can be decomposed as $I(z;s | y) = H(s | y) - H(s | z, y)$, and that $H(s | y)$ is constant with respect to all model parameters, so that the adversary's loss provides a lower bound on the mutual information:
\begin{align}
    I(z; s | y) & = H(s | y) - H(s | z, y) \\
    & \ge H(s | y) - \mathcal{L}_{\textsc{CE}}(\theta, \psi).
\end{align}
We can substitute this bound for the mutual information term in our regularized training objective for the encoder and classifier models:
\begin{align}
    \theta^*, \phi^* & = \argmin_{\theta, \phi} R(\theta, \phi) + \lambda I(z;s | y) \\
    & \approx \argmin_{\theta, \phi} R(\theta, \phi) + \lambda \mathcal{L}_{\textsc{CE}}(\theta, \psi).
\end{align}

We will compute terms for items drawn from each class and collect these to get an overall loss; we weight each term in using inverse class frequencies, which corresponds to enforcing a uniform class prior, and accounts for the possibility of class imbalance in our batching procedure.
This results in the censoring procedure described in Alg.~\ref{alg:adv_conditional}.

\begin{algorithm}[h]
\DontPrintSemicolon
\KwIn{Samples $\{(x_i, y_i, s_i)\}_{i=1}^N$, Encoder $f_\theta$, Adversarial Classifier $\alpha_\psi$, No. classes $C$}
\KwOut{Mutual Information penalty}
$\mathcal{L}_{\textrm{total}} \gets 0$ \;
\For{$i$ in $1 \ldots N$}{
    $z_i \gets f_\theta(x_i)$ \;
}
\For{$c$ in $1 \ldots C$}{
    $\mathcal{S}_c \gets \{z_i : y_i =c \}$ \tcp*[r]{Class $c$ subset}
    \For{$i$ in $1 \ldots N$ where $y_i = c$}{
        $z_i \gets f_\theta(x_i)$ \tcp*[r]{Encode raw data}
        $q_\psi(s_i | z_i, y_i) \gets \alpha_\psi(z_i, y_i)$ \tcp*[r]{Predict subj}
        add $\frac{1}{| \mathcal{S}_c|} \lossCE( q_\psi(s_i | z_i, y_i), s_i )$ to $\mathcal{L}_{\textrm{total}} $ \;
    }
}
\Return $\mathcal{L}_{\textrm{total}}$ \;
\caption{Conditional Adversarial Censoring}
\label{alg:adv_conditional}
\end{algorithm}

\paragraph{Complementary Mutual Information} To perform complementary censoring using an adversarial classifier, we partition the latent space into halves $z = (z^{(1)}, z^{(2)})$. Then, we train two separate adversary models. One tries to reduce the mutual information between the nuisance variable and the first half $I(s; z^{(1)})$, while the other tries to increase the mutual information between the nuisance variable and the second half $I(s; z^{(2)})$. 
\begin{align}
    \theta^*, \phi^* & = \argmin_{\theta, \phi} R(\theta, \phi) + \lambda( I(z^{(1)}; s) - I(z^{(2)}; s)).
\end{align}
The derivation of each of the mutual information terms used in this objective follows the same reasoning as in the case of marginal mutual information; we simply flip the sign of the second term so that minimizing the objective will result in increasing that mutual information.
The resulting censoring procedure described in Alg.~\ref{alg:adv_complementary}.

\begin{algorithm}[h]
\DontPrintSemicolon
\KwIn{Samples $\{(x_i, y_i, s_i)\}_{i=1}^N$, Encoder $f_\theta$, Adversarial Classifier $\alpha_\psi$}
\KwOut{Mutual Information penalty}
$\mathcal{L}_{\textrm{total}} \gets 0$ \;
\For{$i$ in $1 \ldots N$}{
    $(z_i^{(1)}, z_i^{(2)}) \gets f_\theta(x_i)$ \tcp*[r]{Split latent representation}
    $q_\psi(s_i | z_i^{(1)}, y_i) \gets \alpha_\psi(z_i^{(1)}, y_i)$ \tcp*[r]{Predict subj from each half}
    $q_\psi(s_i | z_i^{(2)}, y_i) \gets \alpha_\psi(z_i^{(2)}, y_i)$ \;
    add $\lossCE( q_\psi(s_i | z_i^{(1)}), s_i )$ to $\mathcal{L}_{\textrm{total}} $ \;
    subtract $\lossCE( q_\psi(s_i | z_i^{(2)}), s_i )$ from $\mathcal{L}_{\textrm{total}} $ \;
}
\Return $\mathcal{L}_{\textrm{total}}$ \;
\caption{Complementary Adversarial Censoring}
\label{alg:adv_complementary}
\end{algorithm}

\subsubsection{Mutual Information Gradient Estimation (MIGE) Censoring} \label{sec:apx_mige}
Rather than estimating mutual information or a surrogate quantity, we can use MIGE~\cite{mige} to directly estimate the gradient of mutual information with respect to our model parameters.

\paragraph{Marginal Mutual Information} Consider our encoder model $f_\theta( \cdot )$, and recall that we can sample from the implicit pushforwad distribution $\qt(z, y, s)$ by sampling a tuple $(x, y, s) \sim p_{data}$ from the data distribution and finding its representation $z = f_\theta(x)$. MIGE estimates the gradient of the mutual information $I(z; s)$ as follows (see Appendix A of~\cite{mige} for derivation details):
\begin{align}
    \nabla_\theta I(z; s) 
    & = \nabla_\theta H(z) - \nabla_\theta H(z | s) \\
    \nabla_\theta H(z) 
    & = - \E_{\qt(z, y, s)} \left[ \nabla_z \log \qt(z) \nabla_\theta f_\theta(x) \right] \\
    \nabla_\theta H(z | s)
    & = - \frac{1}{M} \sum_{m=1}^M \E_{\qt(z, y, s)} \left[ \nabla_z \log \qt(z | s) \nabla_\theta f_\theta(x) | s = m \right]
\end{align}
This naturally leads us to the censoring procedure described in Alg.~\ref{alg:mige_marginal}.
Note that in this algorithm, we scale the per-subject terms down by a factor of $1/M$ so that the entropy gradient term $\nabla_\theta H(z|s)$ stays appropriately scaled relative to $\nabla_\theta H(z)$.

\newcommand{\Gz}[0]{\ensuremath{\nabla_\theta H(z)}}
\newcommand{\Gzs}[0]{\ensuremath{\nabla_\theta H(z|s)}}
\begin{algorithm}[h]
\DontPrintSemicolon
\KwIn{Samples $\{(x_i, y_i, s_i)\}_{i=1}^N$, Encoder $f_\theta$, No. nuisance values $M$, Score estimator $F_{score}$}
\KwOut{Gradient of MI}
\SetKwProg{Fn}{Subroutine}{:}{}
\SetKwFunction{Fent}{$Est_{\nabla H}$}
\Fn{\Fent{vectors $\{ z_i \}_{i=1}^T$}}{
    $\nabla_\theta H \gets 0$ ; fit $F_{score}$ to $\{ z_i \}$ \;
    \For{$i$ in $1 \ldots T$}{
        $r \gets F_{score}(z_i)$ \tcp*[r]{Evaluate Score}
        add $r \cdot \nabla_\theta z_i$ to $\nabla_\theta H$ \;
    }
    \Return $ \frac{1}{T} \nabla_\theta H$ \;
}
\;
\For{$i$ in $1 \ldots N$}{$z_i \gets f_\theta(x_i)$}
$\Gz \gets $ \Fent{$ \{ z_i \} $} \;
\For{$m$ in $1 \ldots M$}{
    $\mathcal{S}_m \gets \{ z_i : s_i = m \}$ \;
    add $\frac{1}{M} $ \Fent{$ \mathcal{S}_m $} to $\Gzs$
}

\Return $\Gz - \Gzs$ \;
\caption{Marginal MIGE Censoring}
\label{alg:mige_marginal}
\end{algorithm}

\paragraph{Conditional Mutual Information} To extend the MIGE technique for estimating the conditional mutual information $I(z;s | y)$, we consider a similar decomposition:
\begin{align}
    \nabla_\theta I(z; s | y) & = \nabla_\theta H(z | y) - \nabla_\theta H(z | s, y) \\
    \nabla_\theta H(z | y) & = 
    - \frac{1}{C} \sum_{c=1}^C  \E_{\qt(z, y, s)} [ \nabla_z \log \qt(z | y) \nabla_\theta f_\theta(x) | y=c ] \label{eqn:mige_cond_1} \\
    \nabla_\theta H(z | s, y) & = 
    - \frac{1}{CM} \sum_{c=1}^C  \sum_{m=1}^M \E_{\qt(z, y, s)} [ \nabla_z \log \qt(z | s, y) \cdot \nabla_\theta f_\theta(x) \big| s=m, y=c ]. \label{eqn:mige_cond_2}
\end{align}

We can see that estimating the class-conditional mutual information only requires us to calculate the sum of terms for each class in a given batch of data.
Similarly to the marginal approach, we scale the per-subject terms down by a factor of $1/M$ to keep the two entropy gradient terms scaled similarly.
Also, as with conditional adversarial censoring, we scale the terms for each class subset by the inverse of the class frequencies which compensates for any possible imbalance in our batching procedure; this corresponds to enforcing a uniform class prior.
This results in the censoring procedure described in Alg.~\ref{alg:mige_conditional}.

\newcommand{\Gzy}[0]{\ensuremath{\nabla_\theta H(z|y)}}
\newcommand{\Gzsy}[0]{\ensuremath{\nabla_\theta H(z|s, y)}}
\begin{algorithm}[h]
\DontPrintSemicolon
\KwIn{Samples $\{(x_i, y_i, s_i)\}_{i=1}^N$, Encoder $f_\theta$, No. nuisance values $M$, No. classes $C$, Score estimator $F_{score}$}
\KwOut{Gradient of Conditional MI}
\SetKwProg{Fn}{Subroutine}{:}{}
\SetKwFunction{Fent}{$Est_{\nabla H}$}
\Fn{\Fent{vectors $\{ z_i \}_{i=1}^T$}}{
    $\nabla_\theta H \gets 0$ ; fit $F_{score}$ to $\{ z_i \}$ \;
    \For{$i$ in $1 \ldots T$}{
        $r \gets F_{score}(z_i)$ \tcp*[r]{Evaluate Score}
        add $r \cdot \nabla_\theta z_i$ to $\nabla_\theta H$ \;
    }
    \Return $\frac{1}{T} \nabla_\theta H$ \;
}
\;

\For{$i$ in $1 \ldots N$}{
    $z_i \gets f_\theta(x_i)$
}
$\Gzy \gets 0$ \;
\For{$c$ in $1 \ldots C$}{
    $\mathcal{S}_c \gets \{ z_i : y_i = c \}$ \;
    add  $\frac{1}{ |\mathcal{S}_c|}$ \Fent{$ \mathcal{S}_c $} to $\Gzy$ \;
    \For{$m$ in $1 \ldots M$}{
        $\mathcal{S}_{c,m} \gets \{z_i \in \mathcal{S}_c : s_i = m \}$ \;
        add $\frac{1}{M \cdot |\mathcal{S}_{c}|}$ \Fent{$ \mathcal{S}_{c,m}$} to $\Gzsy$ \;
    }
}
\Return $\Gzy - \Gzsy$ \;
\caption{Conditional MIGE Censoring}
\label{alg:mige_conditional}
\end{algorithm}

\paragraph{Complementary Mutual Information} In the case of complementary censoring, the procedure follows by close analogy to the case of Marginal MIGE censoring.
However, we compute two separate gradient terms - one term $\nabla_\theta I(z^{(1)}; s)$ is used to censor the first half of the latent representation, while the other term $-\nabla_\theta I(z^{(2)}; s)$ is used to force the second half of the latent representation to contain high mutual information with the subject ID.
This results in the censoring procedure described by Alg.~\ref{alg:mige_complementary}.

\newcommand{\GzOne}[0]{\ensuremath{\nabla_\theta H(z^{(1)})}}
\newcommand{\GzsOne}[0]{\ensuremath{\nabla_\theta H(z^{(1)}|s)}}
\newcommand{\GzTwo}[0]{\ensuremath{\nabla_\theta H(z^{(2)})}}
\newcommand{\GzsTwo}[0]{\ensuremath{\nabla_\theta H(z^{(2)}|s)}}
%
\begin{algorithm}[h]
\DontPrintSemicolon
\KwIn{Samples $\{(x_i, y_i, s_i)\}_{i=1}^N$, Encoder $f_\theta$, No. nuisance values $M$, Score estimator $F_{score}$}
\KwOut{Positive and Negative Gradients of MI}
\SetKwProg{Fn}{Subroutine}{:}{}
\SetKwFunction{Fent}{$Est_{\nabla H}$}
\Fn{\Fent{vectors $\{ z_i \}_{i=1}^T$}}{
    $\nabla_\theta H \gets 0$ ; fit $F_{score}$ to $\{ z_i \}$ \;
    \For{$i$ in $1 \ldots T$}{
        $r \gets F_{score}(z_i)$ \tcp*[r]{Evaluate Score}
        add $r \cdot \nabla_\theta z_i$ to $\nabla_\theta H$ \;
    }
    \Return $\frac{1}{T} \nabla_\theta H$ \;
}
\;
\For{$i$ in $1 \ldots N$}{
    $(z^{(1)}, z^{(2)}) \gets f_\theta(x_i)$ \tcp*[r]{Split encoding}
}
$\GzOne \gets $\Fent{$\{z_i^{(1)} \}_{i=1}^N$} ; 
$\GzTwo \gets $\Fent{$\{z_i^{(2)} \}_{i=1}^N$} \;
$\GzsOne \gets 0$ ; $\GzsTwo \gets 0$ \;
\For{$m$ in $1 \ldots M$}{
    $\mathcal{S}^1 \gets \{z_i^{(1)} : s_i = m\}$ ; $\mathcal{S}^2 \gets \{z_i^{(2)} : s_i = m\}$ \;
    add $\frac{1}{M}$ \Fent{$\mathcal{S}^1$} to $\GzsOne$ \;
    add $\frac{1}{M}$ \Fent{$\mathcal{S}^2$} to $\GzsTwo$
}

\Return $\GzOne - \GzsOne$, $\GzsTwo - \GzTwo$
\caption{Complementary MIGE Censoring}
\label{alg:mige_complementary}
\end{algorithm}

\paragraph{Score Function Estimation Details} \label{sec:apx_score_estimation}
\textcite{kscore} frame the problem of score function estimation as a regularized vector regression problem. They use this perspective to provide an improved out-of-sample extension for the previously developed Stein gradient estimator~\cite{li2017gradient}. They also consider the standard Tikhonov regularization approach under this perspective, and give a resulting score estimator. These authors also consider iterative procedures for solving the regression problems that they formulate, including the $\nu$-method of \textcite{engl1996regularization}.
We explore the original SSGE score estimator, as well as the Stein, Tikhonov, and $\nu$-method estimators.
In general, these estimators each introduce a set of hyperparameters that must be tuned, but we find that the Stein estimator performs relatively well across a range of hyperparameters, and frequently performs best among the estimators during our hyperparameter tuning experiments. We note that the score function estimation step of MIGE may be an opportunity for future experimentation, e.g. considering other methods from the energy-based modeling literature (e.g. see \textcite{song2021train} for review).

\paragraph{Adaptive Kernel Length Scales} \label{sec:apx_adaptive_lengthscale} The score function estimators we consider all require computing a kernel matrix for a batch of data points, and in turn these kernels required selecting a length scale.
Since we are estimating pairwise distances in the latent space of a complex non-linear function (i.e. a deep neural network), we cannot estimate a reasonable length scale based on statistics of the dataset alone.
Furthermore, the latent space is changing throughout training, such that a fixed kernel length scale may not be suitable.
Thus, we also consider two adaptive methods of setting the kernel length scale.

First, we consider a batchwise-adaptive length scale using the median heuristic~\cite{sriperumbudur2009kernel}. For a given batch of samples, we simply sets the length scale as the median pairwise distance within the batch.

Second, we consider a pointwise-adaptive length scale inspired by the approach used in t-SNE~\cite{tsne}.
t-SNE embeds a set of high-dimensional points into a lower dimension in such a way as to best preserve the pairwise edge weights of the original point cloud. 
For each point, the edge weight to its neighbors in the high-dimensional (input) space is computed as the conditional probability of a Gaussian centered at the point of interest and evaluated at the neighbor point. The Gaussian variance is tuned by binary search until the distribution over neighbors has perplexity close to a pre-specified value.
Note that this results in non-symmetric pairwise distances, which is not immediately suitable for constructing a kernel matrix, since a valid kernel matrix must be symmetric positive semi-definite (i.e. our kernel matrix $K$ must satisfy $K = K^T$ and $\| z^T K z \| \ge 0, \forall z$).
Thus we adjust this ``fixed perplexity" strategy slightly as follows.
Let the squared L2 distance between points $i$ and $j$ be $d_{ij} = d_{ji} = \| z_i - z_j \|^2$.
For each point $i$, we then find the (non-symmetric) inverse length scale $\beta_i$ using binary search, such that the resulting perplexity of its distribution over neighbors is within a tolerance $1e-5$ of our target perplexity (or up to $100$ steps).
This gives an asymmetric matrix of directed edge weights $\tilde{K}$ having entries of the form $\tilde{K}_{ij} = \exp(-\beta_{i} d_{ij})$.
Then, for a given pair of feature vectors $z_i, z_j$, we compute the desired entry in our final symmetric kernel matrix $K_{ij} = K_{ji}$ by using the average length scale $\beta_{ij} = (\beta_{i} + \beta_{j}) / 2$.
This corresponds to setting the kernel inner product to be the log-space average of the two original values $\tilde{K}_{ij}$ and $\tilde{K}_{ji}$ from the asymmetric kernel matrix $\tilde{K}$:
\begin{align}
    K_{ij} = \exp \left(- d_{ij} \cdot \frac{\beta_{i} + \beta_{j}}{2} \right)
    = \exp\left(\frac{ \log \tilde{K}_{ij} + \log \tilde{K}_{ji} }{ 2 }\right)
\end{align}
\subsection{Divergence Estimation Methods}
Recall from Table~\ref{tab:penalties} and Sec.~\ref{sec:divergence_methods} that we can use some estimate of divergence between $\qt(z)$ and $\qt(z | s)$ as a proxy measure for testing independence between $z$ and $s$. 
By the definition of conditional probability, these distributions will be equal when $z$ is independent of $s$. Thus, we can minimize some divergence measure between these distributions to force these variables to be independent.
In the case of conditional censoring, we can instead compare the distributions $\qt(z|y)$ and $\qt(z|s, y)$ using the same techniques.
In the case of complementary censoring, we can minimize a divergence measure between $\qt(z^{(1)})$ and $\qt(z^{(1)}|s)$, while maximizing a divergence measure between $\qt(z^{(2)})$ and $\qt(z^{(2)}|s)$.

\subsubsection{Maximum Mean Discrepancy (MMD) Censoring} \label{sec:apx_mmd}
The MMD~\cite{mmd} provides one method for computing a suitable divergence measure between a pair of distributions.
\textcite{mmd} show how to compute an unbiased empirical estimate of the squared MMD for the class of functions $\mathcal{F}$ being a unit ball in a universal reproducing kernel Hilbert space, by using a radial basis function (RBF) kernel. 
Given a batch of $N$ samples $\{x_i\}_{i=1}^N \sim p(x)$ and a batch of $M$ samples $\{y_i\}_{i=1}^M \sim q(y)$ this estimate is computed as:
\begin{align}
    & \textrm{MMD}_{\textrm{est}}^2(\{x_i\}_{i=1}^N, \{y_i\}_{i=1}^M) 
    = \frac{1}{N(N-1)} \sum_{\substack{i,j=1 \\ i \ne j}}^N k(x_i, x_j) \nonumber \\
    & + \frac{1}{M(M-1)} \sum_{\substack{i,j=1 \\ i \ne j}}^M k(y_i, y_j)
    - \frac{2}{M N} \sum_{i,j=1}^N k(x_i, y_j). \label{eqn:mmd}
\end{align}
\paragraph{Marginal Divergence} 
We can directly use the estimate of squared MMD from Eq.~\ref{eqn:mmd} for this divergence term by averaging the estimate over values of the discrete nuisance variable $s$.
For a given batch of samples, we consider the entire batch of features $z$ as our sample from the first distribution, and consider the subset of features $z_i$ with a specific nuisance value $s_i=r$ as our sample from the second distribution $q$. 
This results in the censoring procedure described in Alg.~\ref{alg:mmd_marginal}.

\begin{algorithm}[h]
\DontPrintSemicolon
\KwIn{Samples $\{(x_i, y_i, s_i)\}_{i=1}^N$, Encoder $f_\theta$, No. nuisance values $M$}
\KwOut{Estimated squared MMD}
\For{$i$ in $1 \ldots N$}{$z_i \gets f_\theta(x_i)$}
$\mathcal{L}_{\textrm{MMD}} \gets 0$ \;
\For{$m$ in $1 \ldots M$}{
    $\mathcal{S}_m \gets \{z_i : s_i = m \}$ \tcp*[r]{Subject $m$ subset}
    $\ell \gets \textrm{MMD}_{\textrm{est}}^2(\{z_i \}, \mathcal{S}_m)$ \tcp*[r]{Eq.~\ref{eqn:mmd}}
    add $\ell$ to $ \mathcal{L}_{\textrm{MMD}}$ 
}
\Return $ \mathcal{L}_{\textrm{MMD}}$
\caption{Marginal MMD Censoring}
\label{alg:mmd_marginal}
\end{algorithm}

\paragraph{Conditional Divergence} We can naturally extend this approach to enforcing conditional independence between $z$ and $s$ given $y$ by measuring the MMD between the distributions $\qt(z|y)$ and $\qt(z|y, s)$. To compute this conditional censoring penalty for a batch of encoded examples, we compute a term for each class-conditional subset of the batch, and average over these terms.
This results in the censoring procedure in Alg.~\ref{alg:mmd_conditional}.
Unlike with conditional censoring using several other methods, weighing the terms for each class subset is not necessary, since the MMD estimate from Eq.~\ref{eqn:mmd} already scales down based on the number of points being considered from each distribution.

\begin{algorithm}[h]
\DontPrintSemicolon
\KwIn{Samples $\{(x_i, y_i, s_i)\}_{i=1}^N$, Encoder $f_\theta$, No. nuisance values $M$, No. class labels $C$}
\KwOut{Estimated squared MMD}
\For{$i$ in $1 \ldots N$}{$z_i \gets f_\theta(x_i)$}
$\mathcal{L}_{\textrm{MMD}} \gets 0$ \;
\For{$c$ in $1 \ldots C$}{
    $\mathcal{S}_c \gets \{z_i : y_i = c\} $ \tcp*[r]{Class $c$ subset}
    \For{$m$ in $1 \ldots M$}{
        $\mathcal{S}_{cm} \gets \{z_i \in \mathcal{S}_c : s_i = m\}$ \tcp*[r]{Class $c$ and subject $m$ subset}
        $\ell \gets \textrm{MMD}_{\textrm{est}}^2(\mathcal{S}_c, \mathcal{S}_{cm})$ \tcp*[r]{Eq.~\ref{eqn:mmd}}
        add $\ell$ to $ \mathcal{L}_{\textrm{MMD}}$ \;
    }
}
\Return $ \mathcal{L}_{\textrm{MMD}}$
\caption{Conditional MMD Censoring}
\label{alg:mmd_conditional}
\end{algorithm}

\paragraph{Complementary Divergence} We can likewise extend this MMD method to approximating the complementary censoring objectives in Table~\ref{tab:penalties} by performing the same MMD computation in parallel for the two halves of the latent space of our encoder model. This procedure is described in Alg.~\ref{alg:mmd_complementary}.

\begin{algorithm}[h]
\DontPrintSemicolon
\KwIn{Samples $\{(x_i, y_i, s_i)\}_{i=1}^N$, Encoder $f_\theta$, No. nuisance values $M$}
\KwOut{Complementary MMD penalties}
\For{$i$ in $1 \ldots N$}{
    $(z_i^{(1)}, z_i^{(2)}) \gets f_\theta(x_i)$ \tcp*[r]{Split encoding}
}
$\mathcal{L}_{\textrm{MMD}}^1 \gets 0$ ;
$\mathcal{L}_{\textrm{MMD}}^2 \gets 0$ \;
\For{$m$ in $1 \ldots M$}{
    $\mathcal{S}_m^{(1)} \gets \{z_i^{(1)} : s_i = m\}$ ; $\mathcal{S}_m^{(2)} \gets \{z_i^{(2)} : s_i = m\} $ \tcp*[r]{Subject $m$ subsets}
    $\ell^{(1)} \gets \textrm{MMD}_{\textrm{est}}^2(\{z_i^{(1)}\}, \mathcal{S}_m^{(1)})$ ; 
    $\ell^{(2)} \gets \textrm{MMD}_{\textrm{est}}^2(\{z_i^{(2)}\}, \mathcal{S}_m^{(2)})$  \tcp*[r]{Eq.~\ref{eqn:mmd}} 
    add $\ell^{(1)}$ to $ \mathcal{L}_{\textrm{MMD}}^1$ ;
    add $\ell^{(2)}$ to $ \mathcal{L}_{\textrm{MMD}}^2$\;
}

\Return $\mathcal{L}_{\textrm{MMD}}^1$, $\mathcal{L}_{\textrm{MMD}}^2$
\caption{Complementary MMD Censoring}
\label{alg:mmd_complementary}
\end{algorithm}

\subsubsection{``Pairwise" Maximum Mean Discrepancy (PairMMD) Censoring} \label{sec:apx_pairmmd}

As mentioned in Section~\ref{sec:pairmmd}, we can also check for independence between $z$ and $s$ by comparing the latent distributions for different subjects, such as $\qt(z|s=s_1)$ and $\qt(z|s=s_2)$. 
In Algorithms~\ref{alg:pairmmd_bernoulli} and~\ref{alg:pairmmd_clique}, we first describe the two methods used for selecting a subset of all such pairwise comparisons.
We perform this subsetting in order to reduce the time complexity of this strategy.

\begin{algorithm}[h]
\SetKwInOut{KwIn}{In}
\SetKwInOut{KwOut}{Out}
\DontPrintSemicolon
\KwIn{Nuisance values $\{1 \ldots M\}$, Fraction $b$}
\KwOut{Selected subset of nuisance pairs}
$\mathcal{S} \gets \{ \}$ \;
\For{$r$ in $1 \ldots M$}{
    \For{$t \ne r$ in $1 \ldots M$}{
        \lIf{$\textrm{Uniform}(0, 1) \le b$}{add $(r, t)$ to $\mathcal{S}$}
    }
}
\Return $\mathcal{S}$
\caption{``Bernoulli" subset selection}
\label{alg:pairmmd_bernoulli}
\end{algorithm}
\begin{algorithm}[h]
\SetKwInOut{KwIn}{In}
\SetKwInOut{KwOut}{Out}
\DontPrintSemicolon
\KwIn{Nuisance values $\{1 \ldots M\}$, Size $d \le M$}
\KwOut{Selected subset of nuisance pairs}
$\mathcal{S} \gets \{ \}$ ; $\mathcal{T} \gets$ first $d$ of $\textrm{RandPerm}([1 \ldots M])$ \;
\For{$r$ in $\mathcal{T}$}{
    \For{$t \ne r$ in $\mathcal{T}$}{
    add $(r, t)$ to $\mathcal{S}$
    }
}
\Return $\mathcal{S}$
\caption{``Clique" subset selection}
\label{alg:pairmmd_clique}
\end{algorithm}

\paragraph{Marginal Divergence} By close analogy to our approach for MMD censoring in Section~\ref{sec:mmd}, we compute a penalty for minimizing the divergence between $\qt(z|s_i)$ and $\qt(z|s_{j \ne i})$.
Minimizing the average of these MMD terms across each subject-specific subset of a batch provides us a quantitative surrogate for increasing the marginal independence between $z$ and $s$.
The resulting algorithm is described in Alg.~\ref{alg:pairmmd_marginal}

\begin{algorithm}[h]
\DontPrintSemicolon
\KwIn{Samples $\{(x_i, y_i, s_i)\}_{i=1}^N$, Encoder $f_\theta$, No. nuisance values $M$}
\KwOut{Estimated squared MMD}
$\mathcal{L}_{\textrm{PairMMD}} \gets 0$ \;
\For{$i$ in $1 \ldots N$}{
    $z_i \gets f_\theta(x_i)$
}
select pairs $\mathcal{S}$ from $\{1 \ldots M\}$ using Alg.~\ref{alg:pairmmd_bernoulli} or~\ref{alg:pairmmd_clique} \;
\For{$(r, t)$ in $\mathcal{S}$}{
    $\mathcal{S}_r \gets \{ z_i : s_i = r \}$ ;
    $\mathcal{S}_t \gets \{ z_i : s_i = t \}$ \tcp*[r]{Subject subsets}
    $\ell \gets \textrm{MMD}_{\textrm{est}}^2(\mathcal{S}_r, \mathcal{S}_t)$ \tcp*[r]{Eq.~\ref{eqn:mmd}} 
    add $\ell$ to $\mathcal{L}_{\textrm{PairMMD}}$ \;
}
\Return $ \mathcal{L}_{\textrm{PairMMD}}$
\caption{Marginal Pairwise MMD Censoring}
\label{alg:pairmmd_marginal}
\end{algorithm}

\paragraph{Conditional Divergence} We can apply this pairwise MMD censoring approach to enforcing the conditional independence $z \perp s | y$ by computing a similar term for each class-conditional subset of a batch. As before, we take a weighted average across classes to account for possible class imbalance in our sample batches. This results in the censoring procedure procedure described in Alg.~\ref{alg:pairmmd_conditional}.

\begin{algorithm}[h]
\DontPrintSemicolon
\KwIn{Samples $\{(x_i, y_i, s_i)\}_{i=1}^N$, Encoder $f_\theta$, No. nuisance values $M$, No. class labels $C$}
\KwOut{Estimated squared MMD}
$\mathcal{L}_{\textrm{PairMMD}} \gets 0$ \;
\For{$i$ in $1 \ldots N$}{
    $z_i \gets f_\theta(x_i)$
}
select pairs $\mathcal{S}$ from $\{1 \ldots M\}$ using Alg.~\ref{alg:pairmmd_bernoulli} or~\ref{alg:pairmmd_clique} \;
\For{$c$ in $1 \ldots C$}{
    \For{$(r, t)$ in $\mathcal{S}$}{
        $\mathcal{S}_{rc} \gets \{z_i :s_i=r, y_i = c \}$ ;
        $\mathcal{S}_{tc} \gets \{z_i :s_i=t, y_i = c \}$ \tcp*[r]{Class and subject subsets}
        $\ell \gets \textrm{MMD}_{\textrm{est}}^2(\mathcal{S}_{rc},\mathcal{S}_{tc})$ \tcp*[r]{Eq.~\ref{eqn:mmd}}
        add $\ell$ to $\mathcal{L}_{\textrm{PairMMD}}$ \; 
    }
}
\Return $ \mathcal{L}_{\textrm{PairMMD}}$
\caption{Conditional Pairwise MMD Censoring}
\label{alg:pairmmd_conditional}
\end{algorithm}

\paragraph{Complementary Divergence} We can likewise use our pairwise MMD censoring approach to provide a divergence measure for optimizing the complementary censoring objective in Table~\ref{tab:penalties}.
We first partition the latent space of our encoder $f_\theta$, and then compute a subset of the pairwise MMD terms as described above. 
In this case, we will minimize one MMD term to enforce independence between part of the latent representation $z^{(1)}$ and $s$, while maximizing another MMD term to force the model to encode subject-specific information in $z^{(2)}$.
This results in the censoring procedure described in Alg.~\ref{alg:pairmmd_complementary}.

\begin{algorithm}[h]
\DontPrintSemicolon
\KwIn{Samples $\{(x_i, y_i, s_i)\}_{i=1}^N$, Encoder $f_\theta$, No. nuisance values $M$, No. class labels $C$}
\KwOut{Estimated squared MMD penalties}
$\mathcal{L}_{\textrm{PairMMD}}^{(1)} \gets 0$ ; 
$\mathcal{L}_{\textrm{PairMMD}}^{(2)} \gets 0$ \;
\For{$i$ in $1 \ldots N$}{
    $(z_i^{(1)}, z_i^{(2)}) \gets f_\theta(x_i)$ \tcp*[r]{Split encoding} 
}
select pairs $\mathcal{S}$ from $\{1 \ldots M\}$ using Alg.~\ref{alg:pairmmd_bernoulli} or~\ref{alg:pairmmd_clique} \;
\For{$(r, t)$ in $\mathcal{S}$}{
    $\mathcal{S}_r^{(1)} \gets \{z_i^{(1)} : s_i = r\}$ ; 
    $\mathcal{S}_t^{(1)} \gets \{z_i^{(1)} : s_i = t\}$ \tcp*[r]{First half subsets}
    $\mathcal{S}_r^{(2)} \gets \{z_i^{(1)} : s_i = r\}$ ;
    $\mathcal{S}_t^{(2)} \gets \{z_i^{(1)} : s_i = t\}$ \tcp*[r]{Second half subsets}
    $\ell^{(1)} \gets \textrm{MMD}_{\textrm{est}}^2(\mathcal{S}_r^{(1)}, \mathcal{S}_t^{(1)})$ ; 
    $\ell^{(2)} \gets \textrm{MMD}_{\textrm{est}}^2(\mathcal{S}_r^{(2)}, \mathcal{S}_t^{(2)})$ \tcp*[r]{Eq.~\ref{eqn:mmd}}
    add $\ell^{(1)}$ to $\mathcal{L}_{\textrm{PairMMD}}^{(1)}$ ; 
    add $\ell^{(2)}$ to $\mathcal{L}_{\textrm{PairMMD}}^{(2)}$ \;
}
\Return $ \mathcal{L}_{\textrm{PairMMD}}^{(1)}$, $\mathcal{L}_{\textrm{PairMMD}}^{(2)}$
\caption{Complementary Pairwise MMD Censoring}
\label{alg:pairmmd_complementary}
\end{algorithm}

\subsubsection{BEGAN Discriminator Censoring} \label{sec:apx_began}

As discussed in Section~\ref{sec:disc}, we can adapt the discriminator from BEGAN~\cite{began} to provide a signal that approximately measures the divergence between the distributions $\qt(z)$ and $\qt(z|s)$.

\paragraph{Marginal Divergence} To apply this technique for marginal censoring, we use the alternating optimization algorithm from \textcite{began}, but substitute the distribution of ``real" data with $\qt(z)$, and substitute the distribution of ``generated" data with $\qt(z | s)$. We compute a loss term in this way for each possible value of the nuisance variable, and average across these values. For discriminator network $D_\psi$ which functions as an autoencoder $D_\psi: z \mapsto \hat{z}$, define the autoencoder loss for a collection of items as $\mathcal{L}_{\textsc{AE}}(\{z_i\}_{i=1}^N, D_\psi) = \frac{1}{N} \sum_i | z_i - D_\psi(z_i) |$.
Then we use train the discriminator alongside our encoder as described in Alg.~\ref{alg:disc_marginal}.
Note that the discriminator and encoder are optimized in separate steps, using the two loss terms returned by this algorithm.
The BEGAN optimization algorithm includes an additional control trade-off coefficient $k$, which controls the relative magnitude of these loss terms to maintain balance between the two models. This control trade-off coefficient is updated during each step of the optimization as well, with a step size of $\beta$. 
Finally, there is a ``diversity" parameter $\gamma$ determining how far the second distribution may deviate from the first. 
Note that we scale the loss terms for $\qt(z | s)$ down by a factor of $M$ (the number of nuisance values) to keep it scaled equivalently to the loss term for $\qt(z)$.
The resulting censoring procedure is described in Alg.~\ref{alg:disc_marginal}.

\newcommand{\Lae}[1]{\ensuremath{\mathcal{L}_{\textsc{AE}}(#1)}}
\newcommand{\Lpz}{\ensuremath{\mathcal{L}_{p(z)}}}
\newcommand{\Lpzs}{\ensuremath{\mathcal{L}_{p(z | s)}}}
\newcommand{\Ldisc}{\ensuremath{\mathcal{L}_{\textrm{Disc}}}}
\newcommand{\Lenc}{\ensuremath{\mathcal{L}_{\textrm{Enc}}}}
\begin{algorithm}[h]
\DontPrintSemicolon
\KwIn{
    Samples $\{(x_i, y_i, s_i)\}_{i=1}^N$,
    Encoder $f_\theta$,
    Discriminator $D_\psi$,
    Nuisance values $M$, 
    Prev control trade-off $k_{prev} \in [0, 1]$, 
    Control rate $\beta$, 
    Control diversity $\gamma$
}
\KwOut{
    Encoder divergence penalty, 
    Discriminator objective, 
    Next control trade-off
}
\For{$i$ in $1 \ldots N$}{
    $z_i \gets f_\theta(x_i)$\;
}
\Lpz $ \gets \Lae{ \{ z_i \} }$ \;
\Lpzs $ \gets 0$ \;
\For{$m$ in $1 \ldots M$}{
    $\mathcal{S}_m \gets \{ z_i : s_i = m \}$ \tcp*[r]{Subject $m$ subset}
    add $ \frac{1}{M} \Lae{\mathcal{S}_m}$ to \Lpzs \;
}
$\Ldisc \gets \Lpz - k_{prev} \cdot \Lpzs$ \; 
$\Lenc \gets \Lpzs $ \;
$k_{next} \gets k_{prev} + \beta \cdot \left(\gamma \cdot \Lpz - \Lpzs \right)$ \;
\Return \Ldisc, \Lenc, $\textrm{clip}(k_{next}, 0, 1)$ \;
\caption{Marginal Censoring using BEGAN Discriminator}
\label{alg:disc_marginal}
\end{algorithm}

\paragraph{Conditional Divergence} To compute the divergence required for conditional censoring, we perform a similar optimization algorithm, but substitute the distribution of ``real" data with $\qt(z | y)$ and substitute the distribution of ``generated" data with $\qt(z | s, y)$. Thus for each task label, we must consider the subset of the batch belonging to each label and compute a separate loss term.
The resulting censoring procedure is described in Alg.~\ref{alg:disc_conditional}.
Note that we again scale the loss terms for $\qt(z | s)$ down by a factor of $M$ (the number of nuisance values) to keep it scaled equivalently to the loss term for $\qt(z)$. Since we define the autoencoder loss as an average over a set of items, we maintain a balance in the penalty for each class without needing to further scale based on the class counts.

\newcommand{\Lpzy}{\ensuremath{\mathcal{L}_{p(z | y)}}}
\newcommand{\Lpzsy}{\ensuremath{\mathcal{L}_{p(z | s, y)}}}
\begin{algorithm}[h]
\DontPrintSemicolon
\KwIn{
    Samples $\{(x_i, y_i, s_i)\}_{i=1}^N$, 
    Encoder $f_\theta$,
    Discriminator $D_\psi$,
    No. nuisance values $M$, 
    No. classes $C$,
    Previous control trade-off $k_{prev} \in [0, 1]$, 
    Control learning rate $\beta$, 
    Control diversity $\gamma$
}
\KwOut{
    Encoder's divergence penalty, 
    Discriminator's objective, 
    Next control trade-off value
}
\For{$i$ in $1 \ldots N$}{
    $z_i \gets f_\theta(x_i)$ \;
}
$\Lpzy \gets 0$ ; $\Lpzsy \gets 0$ \;
\For{$c$ in $1 \ldots C$}{
    $\mathcal{S}_c \gets \{z_i : y_i = c \}$ \tcp*[r]{Class $c$ subset}
    add $\Lae{\mathcal{S}_c} $ to $\Lpzy$ \;
    \For{$m$ in $1 \ldots M$}{
        $ \mathcal{S}_{cm} \gets \{z_i \in \mathcal{S}_c : s_i = m\} $ \tcp*[r]{Class $c$ subject $m$ subset}
        add $\frac{1}{M} \Lae{\mathcal{S}_{cm}}$ to \Lpzsy \;
    }
}
$\Ldisc \gets \Lpzy - k_{prev} \cdot \Lpzsy$ ; $\Lenc \gets \Lpzsy $ \;
$k_{next} \gets k_{prev} + \beta \cdot \left(\gamma \cdot \Lpzy - \Lpzsy \right)$ \;
\Return \Lenc, \Ldisc, $\textrm{clip}(k_{next}, 0, 1)$ \;
\caption{Conditional Censoring using BEGAN Discriminator}
\label{alg:disc_conditional}
\end{algorithm}

\paragraph{Complementary Divergence} We can also extend this BEGAN discriminator approach to divergence estimation for complementary censoring.
Here, we split the latent representation of the encoder model into two halves $(z^{(1)}, z^{(2)}) = f_\theta(x)$, and operate in parallel on these two halves using separate discriminators.
We add a penalty to reduce the divergence between $\qt(z^{(1)})$ and $\qt(z^{(1)} | s)$, to enforce independence between the first half of the representation $z^{(1)}$ and the subject ID $s$, while adding a penalty to increase the divergence between $\qt(z^{(2)})$ and $\qt(z^{(2)} | s)$, to force the model to encode subject-specific information in the second half of the representation $z^{(2)}$.
The resulting censoring procedure is described in Alg.~\ref{alg:disc_complementary}.

\newcommand{\LpzOne}{\ensuremath{\mathcal{L}_{p(z^{(1)})}}}
\newcommand{\LpzTwo}{\ensuremath{\mathcal{L}_{p(z^{(2)})}}}
\newcommand{\LpzsOne}{\ensuremath{\mathcal{L}_{p(z^{(1)} | s)}}}
\newcommand{\LpzsTwo}{\ensuremath{\mathcal{L}_{p(z^{(2)} | s)}}}
\begin{algorithm}[h]
\DontPrintSemicolon
\KwIn{
    Samples $\{(x_i, y_i, s_i)\}_{i=1}^N$,
    Encoder $f_\theta$,
    Discriminators $D_{\psi^{(1)}}$, $D_{\psi^{(2)}}$,
    Nuisance values $M$, 
    Prev control trade-offs $k_{prev}^1, k_{prev}^2 \in [0, 1]$, 
    Control rate $\beta$, 
    Control diversity $\gamma$
}
\KwOut{
    Encoder divergence penalty, 
    Discriminator objective, 
    Next control trade-offs
}
\For{$i$ in $1 \ldots N$}{
    $(z_i^{(1)}, z_i^{(2)}) \gets f_\theta(x_i)$ \tcp*[r]{Split encoding} 
}

\LpzOne $ \gets \Lae{z_i^{(1)}}$ ; \LpzTwo $ \gets \Lae{z_i^{(2)}}$ \;
\LpzsOne $ \gets 0$ ; \LpzsTwo $ \gets 0$ \;
\For{$m$ in $1 \ldots M$}{
    $\mathcal{S}_m^{(1)} \gets \{ z_i^{(1)} : s_i = m \} $ ; 
    $\mathcal{S}_m^{(2)} \gets \{ z_i^{(2)} : s_i = m \} $ \tcp*[r]{Subject subsets}
    add $\frac{1}{M} \Lae{\mathcal{S}_m^{(1)}}$ to \LpzsOne \;
    add $\frac{1}{M} \Lae{\mathcal{S}_m^{(2)}}$ to \LpzsTwo \;
}
$\Ldisc \gets \left( \LpzOne - k_{prev}^1 \cdot \LpzsOne \right) + \left( \LpzTwo - k_{prev}^2 \cdot \LpzsTwo \right)$\; 
$\Lenc \gets \LpzsOne + \left( \LpzTwo - k_{prev}^2 \cdot \LpzsTwo \right)$ \;
$k_{next}^1 \gets k_{prev}^1 + \beta \cdot \left(\gamma \cdot \LpzOne - \LpzsOne \right)$ \;
$k_{next}^2 \gets k_{prev}^2 + \beta \cdot \left(\gamma \cdot \LpzTwo - \LpzsTwo \right)$ \;
\Return \Lenc, \Ldisc, $\textrm{clip}(k_{next}^1, 0, 1)$, $\textrm{clip}(k_{next}^2, 0, 1)$ \;
\caption{Complementary Censoring using BEGAN Discriminator}
\label{alg:disc_complementary}
\end{algorithm}

\subsection{Additional Dataset Info} \label{sec:apx_datasets}
Here, we give further details about the datasets used in our experiments.

\paragraph{RSVP} This rapid serial visual presentation (RSVP) dataset~\cite{rsvp-dataset} was collected as part of development of a P300 speller system. $10$ individuals were shown a target letter from a $28$ symbol alphabet, and then queried with a sequence of distinct letters, one of which matches the target letter. For each query letter presentation, a corresponding $500$ millisecond window of EEG is recorded using a $16$ channel dry-electrode EEG system. This results in a binary target vs. non-target classification task.

\paragraph{ErrP} This binary EEG classification dataset~\cite{errp-dataset} explored the effectiveness of confirming with a user during a P300 spelling task and measuring for an Error Potential (ErrP) response.
After several standard queries to collect letter evidence, the user is then presented with the best candidate letter inferred by the signal model. The user may then produce an error potential in the case that this letter is incorrect.
Data was collected from $16$ individuals using a $56$-channel dry-electrode EEG system at 600 Hz sampling rate.

\paragraph{ASL} In this American Sign Language (ASL) dataset~\cite{asl-dataset}, $10$ subjects were asked to perform one of $33$ gestures using the right hand, following a visual prompt. Each gesture is performed $3$ times per subject, while EMG data from $16$ muscles of the hand and arm is sampled at $1$ KHz.

\paragraph{ECoG-faces} In the ``Faces-Basic" portion of this electrocorticography (ECoG) dataset~\cite{ecog-faces-basic}, $14$ subjects were shown a random image of either a face or house, while measured with $31$ brain-surface electrodes sampling at $1$ Khz for a duration of $400$ milliseconds per trial.

\end{document}